\documentclass[10pt]{extarticle}

\usepackage{amsmath, amssymb}
\usepackage{authblk}
\usepackage{setspace}
\usepackage{microtype}
\usepackage{longtable}

\usepackage[a4paper, margin=1in]{geometry}

\usepackage{graphicx}
\usepackage{float}
\usepackage{caption}
\usepackage{subcaption}

\usepackage{booktabs}
\usepackage{array}
\usepackage{tabularx}
\usepackage{multirow}
\usepackage{siunitx}

\usepackage{natbib} 
\usepackage[hidelinks]{hyperref}

\title{The Table of Media Bias Elements:\\
A sentence-level taxonomy of media bias types and propaganda techniques}

\author[1]{Tim Menzner}
\author[2]{Jochen L. Leidner}

\affil[1]{Center for Responsible Artificial Intelligence (CRAI), 
Coburg University of Applied Sciences and Arts}
\affil[2]{Center for Responsible Artificial Intelligence (CRAI),
and University of Sheffield, Department of Computer Science}

\begin{document}

\maketitle

\begin{doublespace}

\begin{abstract}
Public debates about “left-” or “right-wing” news overlook the fact that bias is usually conveyed by concrete linguistic manoeuvres that transcend any single political spectrum.  We therefore shift the focus from \emph{where} an outlet allegedly stands to \emph{how} partiality is expressed in individual sentences.  Drawing on 26,464 sentences collected from newsroom corpora, user submissions and our own browsing, we iteratively combine close-reading, interdisciplinary theory and pilot annotation to derive a fine-grained, sentence-level taxonomy of media bias and propaganda.

The result is a two-tier schema comprising 38 elementary bias types, arranged in six functional families and visualised as a “table of media-bias elements”.  For each type we supply a definition, real-world examples, cognitive and societal drivers, and guidance for recognition.  A quantitative survey of a random 155-sentence sample illustrates prevalence differences, while a cross-walk to the best-known NLP and communication-science taxonomies reveals substantial coverage gains and reduced ambiguity.

\end{abstract}


\section{Introduction}

In early summer 2025, a social media trend gained significant traction: Summarized under the umbrella catchphrase ``Propaganda I’m not falling for'', users shared videos of themselves set to pop music, with a list of things they consider propaganda they do not believe \citep{propagandaNotFallingFor2025}.

These lists included all kinds of things, from aesthetics like the ``clean girl look'', over lifestyle trends like the rise in matcha popularity, societal expectations like working a 9 to 5 job, political ideas like ``trickle down economics'', to celebrities and what they apparently represent, like Gracie Abrams. Some of these posts cover a wide range of topics, often reflecting whatever concerns the users posting them, a disproportionate number of whom are young women, in their daily lives. Some are more focused on single issues, from AI over climate change, creative writing and parenting advice to B2B sales, there is rarely a topic not covered. Some are highly ironic, some express deeply serious sentiments. Some users are convinced that ``birth control'' is propaganda they don't fall for, some explicitly include ``being anti birth control'' in their lists.

Looking at this mixed bag of everything that can apparently be considered propaganda, it seems hard to decipher what the term even means as defined by this trend. There is, however, one thing that all of these examples have in common, no matter what they are dealing with or who posted them, from which perspective: they are something the respective user does not like. And they do not fall for it because they do not like it.

This is not just an anecdotal observation based on a social media trend, this notion that for most people propaganda basically just means ``something I don't believe but others wrongly do'', is backed by quite some empirical evidence.

In a Gallup survey from 2020, for instance, 69\% of the respondents expressed concern that the news other people are getting might be biased, while only 29\% were concerned about bias in the news they are following \citep{gallup2020_bias_others_news}.

This result is not an isolated trend, but aligns with the ``third-person effect hypothesis'', which predicts that people think that they themselves are not easily influenced by media, but others are, a hypothesis which could be confirmed in several studies over the last decades \citep{Sun2008}.

The widespread belief that one's own opinions and worldviews are the result of rational reflection and ``common sense'', while those who disagree must have been blinded by ideology and propaganda, also helps to explain other, interesting polls.

According to a 2018 survey conducted by BMG Research \citep{bmg2024bbc}, around 40\% of respondents consider the UK's BBC to be a biased news source. However, there is no consensus among this group about the direction of the bias: 22\% perceive it as left-wing, while 18\% believe it leans right-wing. A 2023 poll about the largest German public broadcaster, ARD, revealed almost identical numbers \citep{ndr2024future}.

These paradoxical findings start to make sense when considering the respondents' self-identified political orientation: Among respondents who identify as ``left-wing'', 41\% perceive the BBC as having a right-wing bias, while only 9\% view it as left-wing. A similar pattern emerges among right-wing respondents: 50\% believe the BBC has a left-wing bias, compared to just 14\% who see it as right-wing.

Circling back to the social media trend, respondents here, just like the social media users, perceive media bias as a matter of positions.

And positions, e.g. a left-right or a Democrat-Republican spectrum on a two-dimensional scale, are often the main framework under which bias is analyzed in research \citep{groseclose2005,solaiman2025biaslab}, as well as by services aimed at providing users with an overview of bias in current news coverage \citep{mediabiasfactcheck2024,otero2021,allsides2024}.

However, when the answer to the question ``how biased is this piece of media?'' on a simple two-dimensional scale appears so highly dependent on the respondent's own position on the scale, how useful is such a framework when discussing media bias? You can either like matcha or not, find the clean girl look aesthetic or not, consider birth control progress or a mistake. Who would decide what the unbiased option here would be?

We believe that instead of looking at media bias and propaganda through the lens of individual topics and their relation to constantly evolving ideologies, the employed techniques and linguistic features when discussing these topics are the key to decide what is bias, what is propaganda (we will discuss a distinction between these often synonymously used terms later).

This work therefore proposes a more complex, granular framework in the form of a taxonomy, focusing on individual, concrete ways how bias can show itself in media, indicated by linguistic features or patterns in reasoning at the sentence level, hoping it to prove more useful when discussing bias than a two-dimensional scale. 

By focusing on the mechanisms of bias, rather than simply labeling its direction, we hope that people are able to reflect on media bias with less influence from their own existing biases. In fact, the approach of providing information about an article's position on the left-right spectrum to encourage exposure to opposing viewpoints can backfire and actually have the opposite effect \citep{ToLabelOrNot}. Readers may use these labels as filters to avoid articles that present perspectives differing from their own. Similarly, focusing on political labels such as 'center,' 'left,' or 'right' can undermine positive effects of media literacy interventions \citep{Spinde2025}.

We believe that avoiding explicit labeling in favor of a more nuanced discussion about bias could foster a more open-minded and unprejudiced approach to the topic.

Second, a type-based approach is more universal. What is perceived as left or right, or which two sides exist for an issue in general, may differ from society to society and from culture to culture. For instance, while advocating for universal healthcare is seen as a left-wing position in the United States, it is supported across mainstream conservative parties in many European countries. In some countries, the left-right spectrum may not even be applicable in the same way it is in the West. Moving away from these categories could make discussions of media bias more universally relevant.

Third, we believe that bias is such a multifaceted phenomenon that reducing it to any simple two-dimensional scale does not do justice to its complexity. 

Ultimately, we think that answers to the question ``how biased is this piece of media?'' become much more interesting and meaningful when freed from the limits of a simple two-dimensional scale and framed within the taxonomy we propose. As the true propaganda does not lie in the matcha, but in the way it is discussed.

\section{Related Work}  \label{label:RelatedWork}
        
\subsection{Media Bias and Propaganda}

Media bias is as old as media itself. From the moment humanity began documenting significant events, these records were shaped by the subjective experiences and perspectives of their creators, as well as the societal context in which they lived. In fact, one of the earliest known artifacts that could be considered a news report, the Victory Stele of Naram-Sin dating back to the 23rd century BC, does not merely describe the Akkadian King Naram-Sin's victory over the Lullubi people but praises it \citep{harvard2016victorystele}.

In fact, the mere idea that journalism could and should be, in any sense of the word, objective and unbiased is a fairly recent invention, arising in the mid-to-late 19th century by news agencies who made it their business model to sell just the facts to newspapers which could then “make their own comments upon the facts" \citep{alden1874washington} and was later formalized and expanded by Journalists like Walter Lippmann \citep{lippmann1920liberty}.

It was also Lippmann, together with his partner Charles Merz and his wife, Faye Albertson Lippmann, who produced the first systematic study of bias in modern newspapers: an analysis of The New York Times’ coverage of the Russian Revolution and Civil War, in which they concluded that “in the large, the news about Russia is a case of seeing not what was, but what men wished to see.” \citep{lippmannmerz1920test}

\subsubsection{What drives media bias and how prevalent is it?}

Several potential sources of bias have been identified. For instance, in a market economy, audience demand is a major driver of media bias. Generally speaking, people prefer news that confirms their worldview, which creates market demand for publications that cater to the biases of their targeted audience and disincentivizes them from challenging these views \citep{GentzkowShapiro2010}. Filter bubbles, which are a major source of bias on social media, are broadly driven by the same mechanism \citep{pariser2011filter}.

As for why this preference exists in the first place, research often explains it with the desire to reduce cognitive load. Being confronted with opinions, perspectives and even facts that contradict one’s default assumptions about the world can be stressful, especially as one’s sense of identity and beliefs are often linked. This discomfort can be avoided by not seriously considering different perspectives in the first place \citep{golman2017information}. This research is often connected to the dual-systems theory, which proposes that humans rely on two distinct modes of thinking: a fast, intuitive system that processes most incoming information and relies heavily on cognitive biases, and a slower, more deliberate analytical system that is used less frequently \citep{kahneman2011thinking}.

Of course, one could describe the relationship between the bias of the media and the bias of its audience as a “chicken–egg” situation, since the biases of the audience, which demand to be confirmed, do not arise spontaneously but are themselves formed in the context of mass media influence. The media is not simply a “victim” of audience biases but actively shapes them, sometimes intentionally, sometimes unintentionally. This occurs, for example, through the biases of journalists working within media organisations, who are often not a representative subset of the population \citep{WJS2022_PersonalBackgrounds}, as well as through the agendas of those who own a given publication \citep{WagnerCollins2014}.

The relationship between these factors can be rather complex. Fox News’ initial decision to confirm Joe Biden’s victory in the 2020 US election, for example, led to a loss of viewers who, arguably in part because Fox News had put great emphasis on conservative talking points alleging the possibility of election fraud before the election, turned to networks promoting the false but demanded narrative that the election was rigged \citep{Khudabukhsh:2022:FNN}. In the end, Fox News caved to market demand and also engaged in spreading falsehoods about the security of voting machines. It was subsequently sued for defamation and made a significant payment as part of a settlement \citep{apnews_fox_dominion_2023}.

Besides attempts to appease and satisfy the audience, sometimes as a direct reaction to accusations of bias \citep{panievsky2022strategic}, the desire to remain in good graces with the subjects of coverage has also been shown to be a relevant factor. Journalists may not want to lose access to interview partners by appearing too critical or may develop personal sympathies through repeated contact \citep{van_der_goot_2021_reporting_on_political_acquaintances}.

Advertising can be considered an aggregation of both factors, since companies and organisations are both potential customers for advertising space in a media outlet and potential objects of coverage. In fact, it has been shown that newspapers tend to report more positively about advertising partners beyond the paid advertisement itself \citep{focke_niessen-ruenzi_ruenzi_2014}.

Moving on from private actors, states and governments also make use of these mechanisms, especially in (more or less) functioning democracies where direct means of pressure or force are more limited \citep{focke_niessen-ruenzi_ruenzi_2014}. Depending on the level of authoritativeness and the strength of independent institutions, other means by which governments can exert influence include withholding or allocating direct funding \citep{enikolopov_petrova_2015_media_capture_empirical} or the use of censorship, either directly enforced or maintained through a culture of self-censorship created by legal or violent pressure on journalists and outlets. Such pressures can also drive bias in cases where a government is not able or willing to protect media creators from threats posed by non state actors \citep{CoE_JournalistsUnderPressure_2017}.

\subsubsection{What are the consequences of exposure to media bias and propaganda?} 

Societies are, ultimately, shaped by the opinions and beliefs that prevail within them, so any measurable effect of media bias and propaganda on the formation of these attitudes has consequences for society as a whole.

It is well established that media bias and outright propaganda can influence voting decisions \citep{AshGallettaPinnaWarshaw2024}, drive polarization \citep{NoyRao2025}, and even leave frequent consumers of such media less informed than those who consume less. This occurs because biased content fosters false narratives while simultaneously increasing individuals’ confidence in these narratives, often more so than among those who are not particularly engaged with politics \citep{Licari2020}.

In a liberal democracy, which relies on the ideal of a well-informed citizenry that considers diverse perspectives, responds to fact-based argumentation when forming political opinions, and remains willing to compromise with those who reach different conclusions, these findings are sobering. This concern becomes even more pronounced when considering the drivers of media bias. When mass-media influence is effective and the actors with the greatest capacity to shape this influence are those who already hold significant power, including governments, influential political organizations, corporations, or wealthy individuals, the result directly undermines another democratic ideal, namely the principle that everyone should have an equal opportunity to have their voice heard and to advocate for their position.

Although these dynamics challenge democracies, at least from a normative perspective, autocracies are almost unimaginable without deliberate efforts to exploit such mechanisms in order to protect their rule. The nature of propaganda varies significantly depending on the level of control a regime can exert. In constrained autocracies, government-aligned propaganda resembles highly partisan publications in democracies and, like them, focuses primarily on persuasion. In totalitarian states, however, the intensity and often absurd character of propaganda serves to reaffirm the state’s all-encompassing power and to discourage opposition \citep{CarterCarter2023}.

The resulting gap between public and private opinion can help protect a regime by hindering the collective organization of discontent \citep{kuran1995private} and by suppressing alternative narratives that might otherwise encourage protest \citep{CarterCarter2021}. It can also bolster perceived support. Yet this equilibrium is far from stable, because new information about actual or perceived expressions of dissent, which suggests that the regime’s grip may not be as firm as previously believed, can rapidly erode that support \citep{BuckleyMarquardtReuterTertytchnaya2023}.

In general, media bias and propaganda appear to be less effective at fundamentally changing minds or creating entirely new attitudes, although autocratic media monopolies may have a relative advantage in this regard \citep{PanShaoXu2021}. Instead, they are more effective at reinforcing existing attitudes, intensifying and radicalizing them, or building on pre-existing belief structures to facilitate the adoption of new ideas \citep{DellaVignaGentzkow2010, SchneiderStrawczynskiValette2025}. There is even evidence that messages contradicting deeply rooted convictions can backfire and further entrench those convictions \citep{PeisakhinRozenas2018}. Nevertheless, for some actors whose goal is not persuasion but polarization, such as states seeking to destabilize an adversary, this outcome, along with the cynicism and erosion of truth that often accompany large-scale propaganda \citep{Shields2021}, may be entirely desirable.

\subsubsection{How can effects of bias and propaganda be mitigated?}

A traditional instrument to counter bias and propaganda, particularly when they are rooted in false information, is the practice of fact-checking. However, there is evidence that, even though it appears to be successful in increasing factual knowledge, especially when paired with correct, alternative explanations \citep{Ecker2022}, repeated exposure to those falsehoods can nonetheless shift attitudes in the intended direction \citep{barrera2020facts}. Still, fact-checks can at least successfully stop or slow the further spread of false information, since prompting users to think about factuality \citep{PennycookEtAl2021} or marking posts as erroneous \citep{ChuaiEtAl2024}, with speed being essential, leads to less sharing of false information.

Besides fact-checking on a case-by-case basis, there is also the generalizable approach of inoculation theory, which postulates that attitudes can be made resistant to persuasion in much the same way the body becomes resistant to disease through medical inoculation, by exposing individuals to weakened persuasive attempts along with refutations in a controlled setting \citep{BanasMiller2013}. We argue that this framework, in particular, would benefit from a comprehensive taxonomy of bias and propaganda types, which could serve as the basis for interventions.

So far, research has shown inoculation to be more promising than fact-checking in fostering medium to long-term resistance to manipulation \citep{BergerKerkhofMindlMuenster2023}. However, the strongest measured effects concern the recognition of these techniques, which does not automatically translate into resistance to them, even though there is evidence that both factors are at least correlated \citep{Ecker2022}.

Since most of these studies are conducted experimentally in more or less controlled environments, it remains unclear how well the effects translate into a long-term ability to spontaneously detect and resist manipulation attempts in real-world settings. Recent studies indicate that success in the first environment does not necessarily transfer to the latter \citep{Wang2025}.

In one way or another, inoculation effects also appear to have a half-life, meaning they may wear down and weaken over time when not repeated. As memories of the intervention fade and individuals are exposed to bias and propaganda again, its effectiveness can diminish \citep{Maertens2025BoosterShots}.

Just as the success of persuasion partly depends on the trust an audience places in the sender of the message and on whether individuals feel they are being manipulated, \citep{PettyCacioppo1979} the success of interventions countering persuasion depends on the perceived credibility of the party conducting the intervention. If people are distrustful and view the intervention itself as an attempt at manipulation, it may even backfire \citep{FransenSmitVerlegh2015}. Therefore, making an intervention appear “political”, for example by framing it as “left” or “right”, instead of focusing on concrete techniques, can reduce its chance of success \citep{Kahan2017}.

In general, successful resistance to bias and propaganda is less linked to classical intelligence and more to an enhanced capacity for critical thinking. This implies that a successful intervention is one that achieves an increase in this ability \citep{Erlich2023}.

There is also increasing research in natural language processing (NLP) aiming to counter bias and propaganda through technical means, either by providing tools that support interventions, for example by automatically highlighting or explaining propaganda techniques as users read news stories, \citep{sharma2025propasafe} or by improving search and recommendation algorithms to show users less biased content by default \citep{MenznerLeidner2025}.

When it comes to technical solutions, we argue that treating bias purely as a technical problem without acknowledging its complex, interdisciplinary, and nuanced nature is insufficient to deliver effective results. Just as social sciences such as political communication and media studies, which have traditionally dealt with the topic, can benefit from integrating NLP methods into their research, NLP researchers should ground their work in these fields’ long-standing insights. However, despite emerging overlap, true interdisciplinarity remains limited. Our taxonomy aims to help bridge this gap. Although it is rooted in the development of automated NLP-based detection methods, it draws heavily on interdisciplinary insights from the aforementioned fields and places all proposed types within this broader context.

\subsection{Attempts to Systematize}

If one adopts a broad historical perspective, the earliest attempts to systematize persuasion techniques can be traced back to antiquity, such as Aristotle’s Rhetoric \citep{aristotle1991rhetoric}. These works focused primarily on spoken discourse directed at an immediate audience, long before the rise of mass media made modern, large-scale propaganda possible. Nevertheless, it can be argued that the core mechanisms of persuasion have remained largely consistent.

With the emergence of mass media, the first efforts to provide an overview of propaganda techniques as a basis for countering such attempts were collected, for example by the Institute for Propaganda Analysis, which published a small taxonomy encompassing the Seven Propaganda Devices in 1937, at a time when totalitarian ideologies that made heavy use of propaganda were spreading in the US and Europe \citep{Schiffrin2022}.

In a similar spirit, many organizations dedicated to promoting media literacy today compile and publish overviews of common techniques outside the scientific literature, like \citep{AllSides_MediaBias}, with a focus on reaching a broader audience rather than on academic discussion.

In academia, while there is a lot of ongoing research on media bias and propaganda in general, efforts to categorize different types in a systematic way have been driven primarily by NLP researchers as a basis for developing detection systems. As of today, the most widely cited work providing a taxonomy of 18 propaganda techniques is \cite{da-san-martino-etal-2019-fine}, later serving as the foundation of the International Workshop on Semantic Evaluation (SemEval) 2020 Task 11: ``Detection of Propaganda Techniques in News Articles'' \citep{da-san-martino-etal-2020-semeval}.

Continuing the historical lineage of propaganda research, their taxonomy incorporates elements from the previously discussed Institute for Propaganda Analysis typology as well as from scholarship on persuasion techniques, further underscoring the interrelations among bias, propaganda, and rhetorical strategies.

A broader, high-level taxonomy, still developed from the perspective of computational detection, was proposed by \cite{Spinde2023MediaBiasTaxonomy}. Drawing on an extensive systematic literature review, they present a coarse-grained framework organized into four main categories, some of which capture phenomena that are difficult to detect without substantial contextual information. Notably, their survey does not reference the well-established taxonomy by \cite{da-san-martino-etal-2019-fine}, discussed earlier. As far as we can determine, this omission stems from the scope of their literature search, which centers on the term media bias and therefore does not surface the adjacent body of NLP work that uses propaganda as its primary keyword, despite the fact that both strands of literature often describe highly overlapping, and in many cases interchangeable, phenomena. This split between two terminological traditions that examine similar underlying mechanisms is a recurring pattern in the field. A recent attempt to bridge both traditions was made by \cite{RodrigoGines-CarrilloDeAlbornoz-Plaza:2024:ExpSysAppl}, another literature review emerging from automated bias detection, which provides an overview of different definitions of media bias and identifies 17 forms that vary according to context and the author’s intention.

In the Terminology chapter, we also elaborate on the relationships between these concepts and clarify how we use them in this work. More broadly, we hope that our contribution helps bridge these two research traditions.

Building on the taxonomy proposed by \cite{Spinde-etal:2023:ArXiv}, \citep{wessel2023mbib} introduce a modified version that is slightly adapted to be more suitable for practical use as a framework for computational, automated bias detection.

While the types in our taxonomy are not directly derived from these approaches or from the other literature discussed in our related work section (see section \ref{label:Methodology} Methodology), we cannot deny that these works have influenced our considerations. We view our work as a continuation of these efforts. Whenever we draw directly on a specific idea or concept, we explicitly reference it in the text.

Our taxonomy differs from the one discussed, as it was developed iteratively with a practical, real-world focus. We worked directly with news examples to shape each category, incorporating interdisciplinary insights into the cognitive drivers behind each bias type as well as their broader societal effects.

Because of this practical foundation, the taxonomy is flexible and can be expanded as new examples and types emerge. It is already more extensive than comparable frameworks, containing 38 types to date. We also provide information on the prevalence of each type based on our data, along with a comprehensive visualization inspired by the periodic table of elements. 

Because our approach is grounded in practical examples, specifically, sentence-level excerpts from news reports, our taxonomy is designed to capture only those forms of bias and propaganda that can be directly observed within short to medium standalone text segments. More indirect phenomena, such as the systematic underrepresentation of certain topics across multiple articles from the same publisher, fall outside the scope of this taxonomy.

To further highlight the similarities and differences between our taxonomy and existing proposals, we provide Table \ref{table:comparison} in the appendix, which presents an attempt to map our proposed types onto those introduced by \cite{da-san-martino-etal-2020-semeval}, \cite{Spinde-etal:2023:ArXiv}, and \cite{RodrigoGines-CarrilloDeAlbornoz-Plaza:2024:ExpSysAppl}. Given the ambiguity inherent in linguistic tasks, individual cases might reasonably be mapped differently. However, the purpose of the table is to demonstrate how our taxonomy both draws on previous research efforts and differs from them in terms of ambiguity, coarseness, completeness, and consistency. We discuss these aspects in greater detail in our Methodology chapter, as they guided our approach to constructing the taxonomy.

\section{Terminology}

Although media bias has been the subject of extensive research for at least several decades, there is no universally accepted definition and, as discussed before in related work, often no clear distinction between media bias and related phenomena like “propaganda” \citep{Hamborg2019-yo}.

A common disagreement is over the role of intention in media bias \citep{Hamborg2019-yo, RodrigoGines-CarrilloDeAlbornoz-Plaza:2024:ExpSysAppl}.
While some authors insist that media bias must, by definition, result from an intentional and conscious decision in reporting, we strongly disagree with that notion. In fact, we argue that, more often than not, biased reporting stems not from a deliberate effort by the writer to deceive their audience but from the unconscious influence of their own biases and the context in which they operate. One might consider themselves unbiased and strive to work accordingly, but as we are all prone to various cognitive biases, biases that can even influence something many would consider objective like our quantitative reasoning capacity \citep{KAHAN_PETERS_DAWSON_SLOVIC_2017}, and must rely on a simplified, subjectivized mental image of the world (``the pseudo-environment,'' as Walter Lippmann called it) \citep{lippmann1922public}, the introduction of bias may be unavoidable.
There is no reason to assume that the individuals contributing significantly to the creation of individual pseudo-environments are themselves exempt from existing within them.
Based on these two factors, that we all possess preexisting narratives about the world in our minds and that we might not even be fully aware of how subjective those narratives are, we define media bias as the tendency to, consciously or unconsciously,
report a news story in a way that supports a pre-existing narrative instead of providing unprejudiced coverage of an issue.

Building on this definition, we describe related terms as sub-phenomena of media bias. Propaganda, for instance, has undergone a remarkable change in connotation over the years, from a mostly neutral term for any attempt to propagate an opinion, including what we would nowadays call “marketing” or “public relations,” to the negatively associated term it is today. However, what has remained constant is that it has always been understood as an intentional effort. So, in the context of this work, we define propaganda as the intentional decision to support a narrative using the types of bias we examine in our taxonomy. 

Not all media bias is propaganda, but all propaganda is media bias (one might even go so far as to call propaganda ``weaponized media bias''). As far as we are concerned, the intentionality behind bias in a piece of media cannot always be easily inferred. Therefore, when discussing types in the context of our taxonomy, one could use both terms interchangeably, since the distinction lies in the intention, not in the technique.


Our analysis and taxonomy primarily focus on media bias apparent in individual, isolated segments of text. We refer to this as “sentence-level” bias, using the term sentence loosely to denote not just a single sentence, but also a small group of sentences that together form a coherent unit.

In contrast, we do not address publication-level bias, which manifests across a larger set of articles from the same source, for example, the over representation of certain topics. Nor do we address article-level bias, which encompasses factors such as the placement of information within an article (e.g., what is emphasized at the beginning versus buried in a middle paragraph) or the selection of accompanying images.

Another “family of terms” frequently used in this field of research includes misinformation, disinformation, and malinformation. While misinformation is commonly defined as any kind of false information shared without intent to deceive, disinformation refers specifically to false information disseminated with the intention to mislead. Both terms are sometimes grouped under the umbrella of “fake news,” although this expression is used more often in informal discourse than in academic research. Malinformation, by contrast, refers to information that is factually correct but shared with the intent to cause harm \citep{wardle_derakhshan_2017}.

We refrain from using these terms because we consider false information to be a sub-phenomenon of media bias (or, in the case of disinformation, propaganda) that is not directly detectable from a sentence alone without additional contextual knowledge about its truth value (see “Unsubstantiated Claims Bias” in our taxonomy). Similarly, we argue that the definition of malinformation is too vague to be of practical use in the context discussed here, especially given that the fight against “fake news” is already invoked as justification for laws explicitly designed to target dissent in autocratic states, where governments claim a monopoly on truth \citep{Mahapatra2024}.

While bias, despite all ambiguity in empirical observation, can at least be formally understood as a deviation from some form of objective reality, the “intent to harm” is entirely dependent on what is considered harmful. Even if agreement existed on what constitutes harm, it would remain unclear whether the relevant intent could be reliably established. Consider pro-democracy activists calling for the end of an autocratic regime: those who value democracy and human rights will regard this as beneficial, while an autocratic regime concerned primarily with its own notion of stability will frame it as harmful. Or consider a newspaper publishing true but classified information about governmental or corporate misconduct leaked by a whistleblower. Even if the whistleblower acted partly for selfish reasons that could be considered an intent to harm, such as revenge or payment by a different actor, does this diminish the value of the information for democratic oversight? Focusing on intended consequences rather than observable forms is not the purpose of this taxonomy. Media bias remains bias even when unintentional, and propaganda remains propaganda even when motivated by seemingly benevolent aims.

For example, the highly successful campaigns promoting polio vaccinations often relied on emotional appeals rather than “cold” factual argumentation \citep{DilawerYasinRasheed2025}.
Aside from those who hold objectively false beliefs, such as claims that vaccines cause autism or other diseases, there is probably broad consensus that reducing, and ideally eliminating, cases of a life-threatening illness that often causes paralysis is a positive outcome. However, neither the intent nor the outcome changes the communicative methods used: emotional appeals still fall under the definition of bias and, when employed deliberately, propaganda. Edward Bernays, in his attempt to rehabilitate the term propaganda, wrote that “The only difference between ‘propaganda’ and ‘education,’ really, is in the point of view. The advocacy of what we believe in is education. The advocacy of what we don’t believe in is propaganda.” \citep{Bernays1928}

As stated in the introduction, we largely agree with this sentiment and believe that approaching the question of bias and propaganda from a perspective centered on moral evaluation is therefore not particularly fruitful. The debate about whether, and under what circumstances, bias and propaganda may be used for positive ends is needed, but separate from the question of establishing when bias and propaganda are present in the first place.


\section{Methodology}  \label{label:Methodology}

The development of this taxonomy is closely tied to the development of \href{www.biasscanner.org}{BiasScanner}\footnote{www.biasscanner.org}\citep{Menzner-Leidner:2025:ECIR}, 
a tool which aims to foster media literacy by automatically highlighting and explaining instances of media bias in news articles. This functionality would not be possible without some kind of structured understanding of what constitutes bias.

For our first experiments\citep{Menzner-Leidner:2024:ECIR}, we relied on the nine  bias types outlined in the Media Bias Identification Benchmark (Linguistic bias, Text-level Context Bias, Reporting-level Context Bias, Cognitive Bias, Hate Speech, Fake News, Racial Bias, Gender Bias, and Political Bias) by \citep{wessel2023mbib}, which is heavily based on the original taxonomy of \citep{Spinde-etal:2021:IPM}. 

However, while analyzing sentences (from their dataset and beyond) using their taxonomy, we increasingly began to notice what we would consider limitations in the context of our goal:

\textbf{Ambiguity:} We consider the boundaries between the categories to be not clearly defined. For instance, ``Linguistic Bias'' is described as encompassing ``all forms of bias induced by lexical features, such as word choice and sentence structure,'' whereas ``Text-level Context Bias'' is defined as involving ``words and statements [that] can shape the context of an article and sway the reader’s perspective.'' From these definitions, it is not clear where, for instance, the usage of euphemisms would fit. 

\textbf{Coarseness:} Naturally, limiting the taxonomy to nine bias types inevitably introduces a degree of coarseness, which may be suboptimal for detailed analysis. Our main critique, however, concerns the large disparity in coarseness across categories. For example, Linguistic Bias, as described above, aims to capture ``all forms of bias induced by lexical features,'' while Discrimination is split into only two much narrower subtypes: ``Gender bias'' and ``Racial bias.''

\textbf{Completeness:} The issue of coarseness is directly tied to the problem of completeness, as some bias types are simply not covered. This is particularly evident in areas where a more granular category would be logical. For instance, because there is no overarching category for all forms of discrimination and only two specific subtypes, there is no way to categorize discrimination based on other criteria, such as religion, sexual orientation, or disability. At the same time other general concepts like fallacies or specific forms of argumentation (e.g Ad Hominem attacks) are completely missing.

\textbf{Inconsistency:} We also view this approach as mixing different layers of bias. While the taxonomy was developed for a dataset consisting of individual sentences, some categories reference bias that cannot be identified from single sentences alone. This is especially the case for ``Cognitive Bias,'' which is defined as the bias introduced by readers’ decisions regarding ``which articles to read and which sources to trust.'' While this indeed constitutes a form of bias, it occurs internally within the reader’s mind and is not directly observable in the media itself, unlike ``Linguistic Bias,'' for example.

When investigating other taxonomies employed for bias classification (see section \ref{label:RelatedWork} Related Work), we found that these points of critique were not unique but rather occurred, in one form or another, quite commonly.

For this reason, we decided to develop our own taxonomy, grounded in the practical examples we encountered during our experiments and applications. 

Based on our original points of critique, we formulated three criteria for determining and differentiating individual types:

\begin{enumerate}

    \item Each bias type must fit our original definition of media bias as supporting a pre-existing narrative over an unprejudiced assessment. 

    \item Each bias type must be identifiable based solely on the content of a single sentence or continuous paragraph, without requiring context from surrounding parts of the article or external sources.

    \item Each bias type must be ``elementary'', as in theoretically able to stand on its own (even though in practice one sentence often belongs to several types) and introduce bias, as defined by 1., independently.

\end{enumerate}

As the basis for identifying types for our taxonomy, we drew on four sources of data:

\textbf{User-collected sentences:} 16,229 sentences flagged as biased by our BiasScanner 
browser add-on or web demo, based on texts submitted to the system by users and collected from 3,710 scans between February 2024 and August 2025. All data is fully anonymized, making it impossible to track individual users’ reading habits. 
The same applies to tracking the origin of texts input by users, including the type of source (news sites, blogs, social media comments, etc.). URLs are only logged when the add-on is used, not the web demo, and only when the user explicitly clicks a button to mark the bias report for the news story they just analyzed as a highlight for us. Consequently, we can make only an educated guess based on the 167 URLs we received through this method, with no way of confirming whether they are truly representative. Slightly more than half of these URLs appear to belong to news article sites, with the five most frequent being, in descending order: tagesschau.de, foxnews.com, bild.de, zeit.de, and bbc.com. The remaining URLs largely link to (political) blogs, websites of parties and organizations, social media comments, and Wikipedia articles.

About 99\% of the sentences were in either German (8,456) or English (7,665). Sentences in other languages could not be considered due to a lack of speakers on our side.

\textbf{A dataset we collected for an experiment of ours:} 9,284 sentences from Reuters and Fox News articles on topics relevant to U.S. political discourse during 2008/2009.\citep{menzner2025biasrank}

\textbf{BABE subset:} 951 sentences pre-labeled as biased, drawn from the BABE dataset, which includes material from 14 U.S. news outlets.

\textbf{Personal observations:} Individual examples encountered during our own personal consumption of news and related content, counted within the 16,229 sentences above  as we submitted them through the add-on as well.

Using this data, our process was as follows:

\textbf{Initial taxonomy:} We began with a bare-bones taxonomy informed by existing taxonomies summarized in the Related Work section above.

\textbf{Sampling:} We drew random subsamples from our data.

\textbf{Manual annotation:} We manually examined drawn sentences to determine whether they could be considered biased. If so, we checked whether the sentence in question fitted to an existing category in our taxonomy.

\textbf{Formulating new types:} For sentences not covered by existing types, we analyzed the nature of the bias by consulting literature and applying our own reasoning. When appropriate, we formulated a new bias type in line with our three criteria. For each bias type, the process is outlined in more detail in the respective discussion. 

\textbf{Refinement:} Throughout this process, we continuously evaluated both the coherence and distinctiveness of our types, aiming to maintain a reasonable balance between the two.

\textbf{Grouping:} Based on observed similarities between different bias types, we organized them into broader groups.

Of course, with this (or ultimately any other) process, it cannot be guaranteed that we have covered all possible ways in which bias could express itself on a sentence level as defined by our three criteria. It is possible that types of bias exist which simply have not yet appeared in our data over the last two years.
Also, for something as highly ambiguous as language and linguistics, the application of the different criteria and their granularity remains, to some extent, a matter of interpretation, where different outcomes could be equally valid. Using the same criteria and the same sentences with the same goal as we did, a taxonomy created by other researchers might show some differences, as they may have decided to merge two of our types into one, or group certain types into a different upper category.

\section{The Table of Bias Elements}

It is for this reason that we decided to borrow the periodic table of elements from chemistry as an analogy for our taxonomy.

Just like the table of elements, constructed as incomplete both at the time of its creation and even today, and continuously evolving as new types are discovered, our taxonomy can and hopefully will be extended with additional types of bias in the future, iteratively approaching the goal of providing a complete overview of all sentence-level media bias types. This analogy also emphasizes the practical nature of our approach, as, similar to the discovery of many elements in the table, new bias types have been added based on practical evidence of their existence through concrete examples.

The structure of the periodic table of elements, where elements with similar characteristics are arranged into groups that are often highlighted by color in visual representations, also served as a model. Arranging our taxonomy in this established way may provide a clear and easily comprehensible framework for understanding the different types of sentence-level bias.

\begin{figure} [!h] \centering \includegraphics[width=0.8\linewidth]{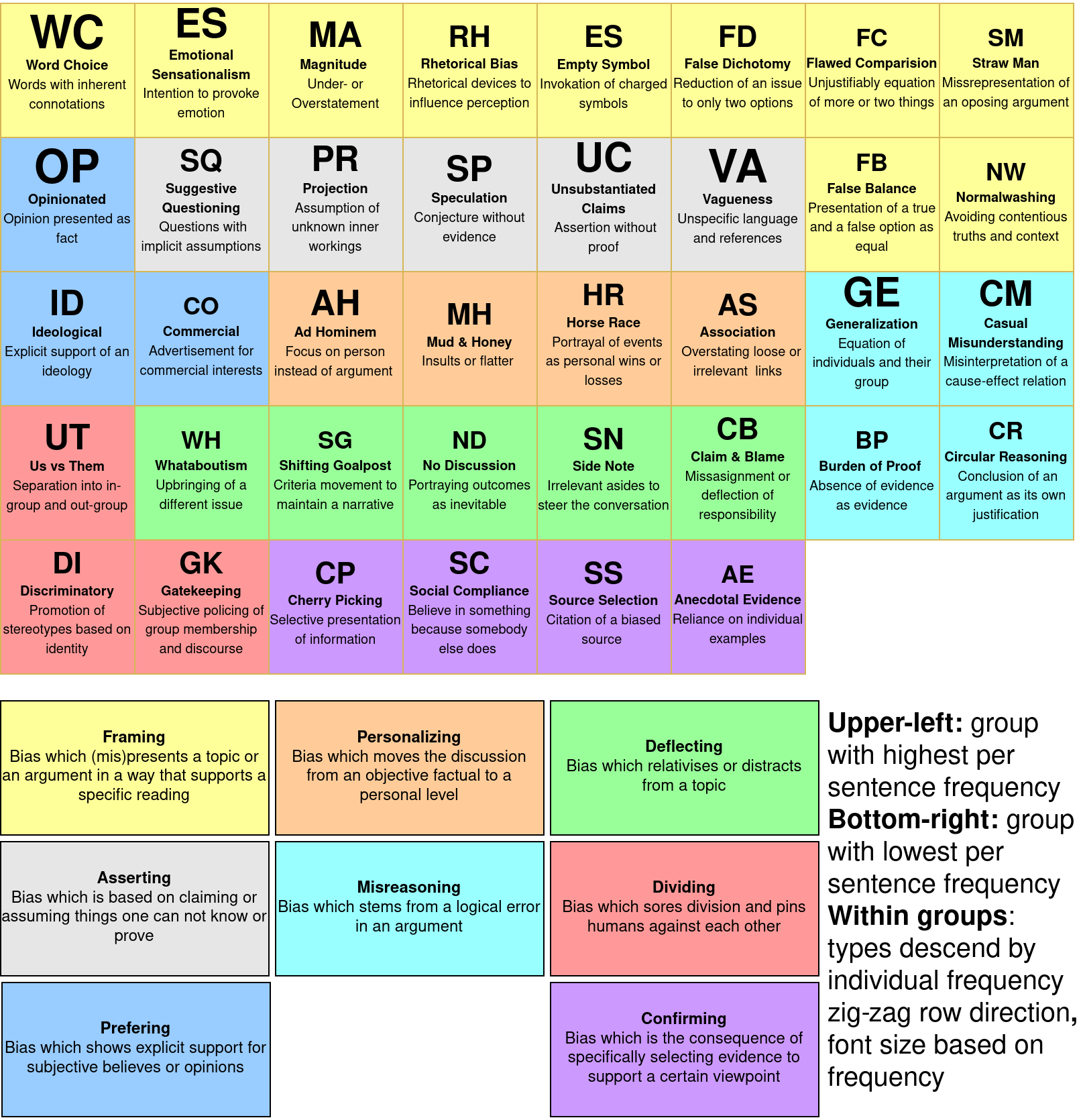} \caption{Our Table Of Bias Elements} \label{fig:bias-taxonomy} \end{figure}

The table of bias types is shown in figure \ref{fig:bias-taxonomy}, and the individual types are further explained below. All examples displayed for the individual types are drawn from the real-world data mentioned above. While these examples may not always illustrate a concept as clearly as the theoretical, constructed examples often used in the literature, such artificial examples rarely reflect how reporting actually looks in practice. We therefore chose to use real-world examples, even though they may not be as straightforward and may include elements of other bias types, since this more accurately represents how biases occur in practice (Our taxonomy explicitly supports multi-type classification). Some of the example sentences given here were originally in German language and have been translated by us for this article.

The table layout for bias groups and their corresponding types was determined by frequency. We randomly sampled 155 biased sentences from our dataset, annotated them using the finished taxonomy, and counted occurrences for each type.

For presentation, the table was designed with 8 × 5 cells. Group placement within the table is determined by the absolute frequency of bias types belonging to this group in our sample. The group with the highest number of occurrences (“Framing”) was placed in the upper-left corner.

Within each group, bias types were ordered by frequency: the most frequent type first (in case of “Framing”, that was ``Word Choice''), followed by the second most frequent, and so on. In the first row, ordering proceeds left to right. Each row contains eight types before breaking to the next line. To keep groups visually connected, the ordering direction alternates: odd rows run left to right, while even rows run right to left. This zig-zag pattern ensures consistent grouping and frequency-based ordering. To enable frequency comparisons between bias types across groups, we introduced five frequency tiers, indicated by decreasing font sizes of the type name: Very high (found in at least 25\% of our sentences), High (15\%–24\%), Medium (8\%–14\%), Low (5\%–7\%), and Very low (1\%–4\%).

For detailed prevalence counts, see Table \ref{tab:bias-absolute-percentages-multicol} in the attachments.

\subsection{Asserting}

Bias types in this category involve claiming or assuming things that cannot be known or proven, in order to support a narrative.




\subsubsection{Projection Bias}

This occurs when thoughts, feelings, motives, or intentions are attributed to others, be it individuals, groups, or entities, without sufficient evidence or direct statements to validate such claims.

\textit{``It feels very much like the writers and creators of the new line of Star Trek shows, which also includes 'Picard,' don’t think just writing a good TV show is important enough work. Like so much of corporate America they think they have to save the country and the planet while they do it.''}

As social beings, it’s natural for humans to constantly question the motives behind others’ words and actions. Yet, without the ability to truly see inside someone’s mind, forcing us to rely on our own, heavily biased internal narratives, this process often reveals more about ourselves than about the people we’re trying to understand.

For an instance, people tend overestimate the rate to which their opinions, beliefs, and behaviors are shared by others \citep{ross1977false}.

Consequently, when confronted with disagreement, especially on topics they are emotionally invested in, they often attribute opposing views to negative motives, not considering the possibility that somebody could come to a different conclusion in good faith \citep{Reeder2005}.

\subsubsection{Speculation Bias}

This is the practice of engaging in speculating based on conjecture about situations or outcomes rather than relying on concrete facts and definitive evidence.

\textit{``Is North Korea sending soldiers to help Putin rebuild the occupied territories? These are still just rumours. But are they really that unlikely?''}


This bias is partly driven by a phenomenon called ``need for closure,'' which describes an individual’s desire for a firm answer to a question and an aversion toward ambiguity \citep{kruglanski1996motivated}.

So it is once again consumer demand which gives news outlets an incentive to engage in speculative coverage when facts are unclear, rather than waiting until the situation is more clear and risk losing part of their audience to a competitor, which pretends to be able to satisfy the need for closure.
It has been shown that need for closure is especially high when threatening topics like terrorism are being discussed \citep{orehek2010need}.
One could also draw a connection to the negativity bias (see ``Emotional Sensationalism Bias'' for a discussion of this phenomenon).

Another potential driver of speculation is overconfidence. Studies have repeatedly shown that humans tend to be more confident in their predictions than is objectively warranted, and that those who appear especially confident are perceived as more competent than those who offer a more realistic assessment of what they know and do not know \citep{Anderson2012StatusEnhancement}.

Another explanation for the prevalence of speculation in the media, especially around these kinds of topics, has been given by Richard Grusin, who argues that, in order to avoid appearing caught off guard when a catastrophe happens, there is an incentive to ``prepare'' for potential disasters in advance by regularly mentioning the possibility of negative outcomes and crises in the reporting \citep{grusin2010premediation}.

For the effects, it has been shown that an increased presence of speculative news also decreases the remembered certainty of unrelated, factual news reports. In other words, when a news outlet engages in speculation, the line between conjecture and confirmed fact becomes blurry across its entire reporting, eroding trust in objectively verifiable information \citep{Brand2023}.



\subsubsection{Suggestive Questioning Bias}

This is the practice of posing suggestive questions that contain implicit assumptions or lead the audience towards a preconceived notion, often used to promulgate subjective beliefs or doubts under the pretense of neutral inquiry. 

\textit{``Is he a left-wing radical or is he a crowd-pleasing moderate? Is he a uniter? Is he actually religious? Is he a socialist? Does he hate America? And, in light of the confidence he projects, why can’t he seem to take criticism?''}

By framing statements as suggestive questions, one can subtly push a narrative while avoiding direct responsibility, claiming to be ``not an expert and just asking questions''. The actual answer is irrelevant since the question itself already implies the ``true answer''. It is particularly effective because it encourages the audience to arrive at the intended conclusion on their own, reinforcing the illusion of independent thought.

The now-famous experiments by Elizabeth F. Loftus showed that suggestive questions can even alter how people remember events they witnessed. After witnessing an event, questions implying a certain detail (whether true or false) often led participants to later recall that detail when asked to remember the event \citep{Loftus1975}.

\subsubsection{Unsubstantiated Claims Bias}

This describes the practice of making claims, statements, or assertions that are substantive enough that they should be supported by evidence, and presenting them as factual without actually providing adequate evidence or references to support their validity.

\textit{``Read all about Dominion and Smartmatic voting companies and you'll soon understand how pervasive this Democrat electoral fraud is, and why there's no way in the world the 2020 Presidential election was either free or fair.''}


This type of bias goes beyond the presentation of an issue, it relies on the spread of outright fabricated information to support the desired narrative, closely aligning with fake news, misinformation, and disinformation. Besides that, spreading false information has also been linked with the goal of generally eroding the concept of an objective truth, making it easier for all kinds of future false narratives to be pushed \citep{Altay2024}.

An important principle driving its viability is the illusory truth effect, the simple repetition of a claim, even without the presentation of any supporting evidence, is enough to increase its perceived truthfulness (interestingly, the second repetition is already the most effective, with diminishing returns for each further iteration) \citep{Hassan2021}.

Even when fact-checked, this principle may still apply, highlighting the importance of discerning not only what to fact-check, but also how to present it without inadvertently legitimizing unfounded claims \citep{barrera2020facts}.

Subsequently, attempting to spread misinformation through serious fact-checking sites has become a strategy of Russian disinformation campaigns. In what has been dubbed ``Operation Overload'', agents posing as ``concerned citizens'' have been contacting those sites, asking if they could check a certain fabricated story, which would otherwise not have gained much visibility in the first place \citep{atanasova2024operation}.

Still, prompting news audiences to reflect on the accuracy of a news story before sharing has been shown to reduce the sharing of false information \citep{pennycook2022accuracy}.

This could be seen as further supporting the notion that the perception of unsubstantiated claims, like potentially other types of bias, is closely linked to the dual process theory. This theory postulates that humans dispose of two different cognitive systems, one being conscious, slow and based on critical and rational thinking, the other one being subconscious, fast and based on stereotypes and learned patterns. In fact, research indicates that people showing higher critical thinking skills also show less likelihood of believing unsubstantiated claims, while those relying on intuition do so more often \citep{bensley2023critical}.

Critical thinking should, however, not be confused with general intelligence, in fact there is evidence to believe that intelligence as traditionally measured does not necessarily translate to a better ability of avoiding bias, as intelligence-associated skills like reasoning capabilities could also be used in favor of convincing oneself of a biased perspective (see motivated reasoning) \citep{stanovich2013myside}.

It should be noted that unsubstantiated claims are not necessarily the same as false claims, despite significant overlap. While it is possible to determine whether evidence for a claim is present or absent, the truth of the claim itself may not be easily established without additional context. Not every unbacked claim in an article might be outright false. In fact, given the high prevalence of claims without explicit sources in news articles, it is reasonable to assume that most are not and linking sources is just not as common in journalism as it could be \citep{reich2023justifying}.

%

\subsubsection{Vagueness Bias}

This is the usage of phrasings so broad and unspecific that they can mean anything and nothing without further clarification, like the usage of catch-all terms (such as ``actions'' or ``factors''), vague references (like ``experts'' or ``the people'') or deliberate concealment of specifics.

\textit{``Experts, entrepreneurs and others have also been quick to condemn the announcement.''}


There is evidence that ambiguity or vagueness can be a powerful persuasive strategy. It's not without reason that this style of language is so common in advertising or political messaging. Experiments show that vague statements are processed more quickly, without the cognitive effort required to identify exact referents. This opens the door to more shallow processing, where gaps are filled with one’s own projections, often aligning with the meaning one would prefer. In this way, a message can be convincing to several people with fundamentally different preferences, in a way that a precise, clear message that inevitably alienates some members of an audience, never could be \citep{mannaioli2024vagueness}.

Additionally, when if an unclear reference is given, there is a connection to the social compliance bias, as the goal, here as there, is to increase the impact of a statement by implying that many other people, or people whose opinions hold particular importance, support it. It is also a great way to feign distance from an argument by masquerading it as someone else's opinion. However, unlike social compliance bias, there does not actually need to be an external source of validation. The claim of support is formulated so broadly and unspecifically that one could insert it in front of an internally manufactured statement. Plus, the vagueness makes it difficult to refute the statement, as there is no clear ``point of attack''.

\subsection{Confirming}

This category is defined by a specific selection of evidence that supports an existing viewpoint.

\subsubsection{Anecdotal Evidence Bias}

This bias stems from relying on individual 
stories or examples rather than considering broader and more representative evidence when forming conclusions. 

\textit{``What was remarkable in 2008 was how quickly Americans abroad sensed a change of mood. On the night of November 4, American expatriates posted jubilant messages to social networking sites like Facebook saying it was cool to be American again.''}

This bias is particularly influential, as it is well established that, for topics with high emotional engagement, anecdotal stories shape people's opinions more effectively than statistical facts \citep{Freling2020}. This is not surprising, as storytelling and imagination have always played a significant role in human development \citep{gottschall2012storytelling}. Dull statistics simply cannot paint the same vivid images in our minds as stories about lived experiences by other humans can.
For the same reason, journalists often construct their articles around anecdotal stories as part of ``narrative storytelling'' to drive engagement \citep{Kriecken2017}.


\subsubsection{Cherry Picking Bias}

This form of bias is evident when news stories give undue prominence to aspects and certain details of a news story that endorses a certain viewpoint, while omitting information that would contest it.

\textit{``Only half of those surveyed considered the political orientation of ARD programmes to be ‘balanced’, while 22 percent rated them as ‘tending to the left.''}

Consider the given example taken from the report of a conservative German newspaper about a poll on the perceived political bias of the public broadcaster ARD. It is a popular conservative talking point to accuse them of being left-wing, so the numbers here are presented in a way that gives the impression that the general public agrees with this sentiment. However, if one sums up the mentioned proportions, it becomes obvious that some information must be missing. And indeed, 19\% of respondents, roughly the same proportion as those who consider ARD left-leaning, perceive it as conservative. While the 22 percent who consider ARD as left even warrant their own subheadline, this piece of information is completely absent from the article, arguably because it doesn’t align with the article’s overall framing. Based on the same critique, the newspaper in question (FAZ) received a reprimand from the German Press Council \citep{presserat2024faz}.)

While cherry-picking facts might be equally misleading as outright false information, it is often perceived far less negatively by readers, making it effective at influencing public opinion, while posing less risk to the communicator’s credibility \citep{Li2024}. 

On the sentence level, this type of bias may not be as easily detectable without additional context about the events described. However, certain indicators can still serve as clues, for example, the complete absence of alternative perspectives or voices, large discrepancies in the space allocated to different viewpoints, missing context for quotations or information and statistics that do not add up.

\subsubsection{Social Compliance Bias}

This bias occurs when an argument is deemed valid or true simply because it is supported by someone with high social standing or personal involvement, an authority figure or because it aligns with the current or traditional beliefs of a group one identifies with, or of a large general group of people.

\textit{``World leader agrees with Vance that mass migration is threat to 'daily life.'''}

It is driven by several powerful social principles. Humans have a natural tendency toward conformity; generally, we don't want to stand out, be isolated, or be in conflict with a group. Instead, we seek to fit in, be accepted, and liked. This principle was famously demonstrated in Solomon Asch's conformity experiments, where participants showed a tendency to agree with a group that two clearly different-length lines were the same length. (Interestingly, just a single dissenter was enough to significantly reduce this effect.) \citep{Asch_1956}

A few decades later, when the experiment was repeated using modern magnetic resonance imaging, it was shown that participants might not have even acted against their better judgment, as the areas of the brain responsible for space perception were more active than those responsible for resolving contradictions, indicating that their brains were trying to distort reality to match the group's opinion \citep{BERNS2005245}.

In a way, this echoes George Orwell’s famous line from 1984: “The Party told you to reject the evidence of your eyes and ears. It was their final, most essential command.” \citep{orwell1984}

Perhaps even more well-known are the experiments of Stanley Milgram, which revealed the dangers of blind obedience to authority figures. His studies showed how easily people could be persuaded to perform unethical acts, like inflicting pain on others through electric shocks, simply because a trusted authority figure told them to \citep{milgram1963behavioral}.

The strong influence of groupthink on beliefs was also demonstrated in experiments where participants’ support for specific policies depended largely on whether the policy was presented as a proposal from their preferred political party or the opposing side. Interestingly, most participants were convinced that it was only the other side, not themselves, who were influenced in this way \citep{Cohen2003PartyOP}.

Like most potentially harmful cognitive biases, however, these tendencies originally served a purpose in the development of human culture. Joseph Henrich, for instance, argues that tradition and conformity can serve as ways to preserve knowledge within a culture. He observed the Tukanoan people in the Amazon lowlands, who peeled, grated, soaked, and boiled manioc multiple times in a labor-intensive process before consuming it. They could not explain why they followed these steps, other than that it was tradition. What they didn't realize was that by doing this, they were reducing cyanide levels to a safe level, thus preventing long-term poisoning \citep{Henrich2015}.

\subsubsection{Source Selection Bias}

This form of bias arises from citing sources that have a high likelihood of being themselves biased regarding the discussed topic, without providing further contextualization.

\textit{``The port of Mariupol in Russian-controlled territory of Ukraine is operating at full capacity, the TASS news agency reported on Wednesday, citing port officials.''}


This might either be an intentional choice, similar to the Cherry Picking Bias, used to support a particular narrative, or simply the result of journalistic negligence, especially when working under time and financial pressure. It is much easier to rely on statements from parties involved in an issue, such as police reports or press releases, which are frequently released to steer the media's coverage of the situation, than to conduct a thorough investigation. 

It has been shown that people trust information less if they perceive it to be from a source they consider biased \citep{wallace2020when}. 

Therefore, it is a logical consequence that actors aiming to push a narrative often resort to astroturfing, a strategy that hides the original messenger, giving the impression that it originates from, and is supported by, independent actors, like regular citizens or groups with a positive image like small business owners \citep{cho2011astroturfing}.




\subsection{Deflecting}

The types in this category are characterized by attempts to deflect discussion, either by relativizing the issues at hand or by steering the conversation elsewhere.

\subsubsection{Claim \& Blame Bias}

This bias occurs when responsibility is rejected or wrongly assigned, whether by giving undue credit, scapegoating, playing the victim, or outright reversing blame onto the actual victim.

\textit{``Instead, the US and its European puppets placed Russia in an impossible position with regards to the ongoing militarization and Nazification of Ukraine, forcing it to respond as any other country concerned about its national security would.''}

Taking the credit for positive achievements while outsourcing the responsibility for perceived failures to external factors appears to be a very natural thing to do, as this disposition has been shown in a wide variety of areas \citep{Mezulis2004}.

Likewise, on a group level, it has been shown that individuals, who are feeling like their group is unfairly accused of wrongdoing like discrimination, often react by claiming victim-hood themselves, partly driven by the  desire to shut down further criticism \citep{Danbold2022}.

These calculations do not seem entirely unfounded, as other studies have demonstrated the effectiveness of tactics such as escaping blame and eliciting sympathy by successfully adopting a victimized stance \citep{GRAY2011516}.

At the same time, research on the so-called DART strategy (Deny, Attack, and Reverse Victim and Offender) has shown that perpetrators of interpersonal violence can successfully lead an audience to attribute part of the blame onto their victims \citep{Harsey13092020}.

Furthermore, studies have highlighted how scapegoating can effectively redirect blame and, at least temporarily, suppress conflict, especially when the scapegoat holds a lower social status than the aggressor \citep{Antonetti2021How}.

\subsubsection{No Discussion Bias}

This denotes the practice of painting the outcome of a matter as inevitable and beyond debate, shutting down disagreement and steering conversation only toward adaptation or acceptance.

\textit{``California's a bellwether state. What happens here, blows east.''}



This bias heavily relies on ``inevitabilism'' \citep{renner2025inevitabilism}, framing what is actually speculation or merely a premise open to argument as if the matter had already been decided. Engaging with such an argument is not possible without first accepting its premise. The conversation thus shifts from asking whether something is true to discussing how to deal with the fact that it is true, even though that would actually still be open to debate.

Research on system justification theory suggests that this can, in fact, be a highly effective way to steer a conversation. Once a premise is accepted as reality, support for that premise tends to increase, even among those who initially opposed it while it was still under debate. This effect is often explained as a coping mechanism for cognitive dissonance: the internal conflict between disliking a new reality and having to accept it can be resolved by rationalizing the reality as legitimate or even beneficial. This rationale is also given as a reason for the paradoxical finding that those disadvantaged by a system are often the most likely to defend it, as they have the greatest psychological need to justify it \citep{jost2003social}.

It is also often visible in the form of the ``thought-terminating cliché,'' as coined by Robert Jay Lifton in his analysis of totalitarian ideologies, referring to the reduction of complex human problems into simplistic, easily memorizable phrases that claim to answer said problems but in reality lack real substance \citep{lifton1961thought}. 
The power of established, simple phrases, like well-known sayings, in persuasion, especially in situations of limited thinking, has also been shown in empirical research \citep{familiar_phrases_peripheral_persuasion_1997}.

\subsubsection{Shifting Goalpost Bias}

This bias occurs when a position is relativized, often in response to criticism, by excluding counterexamples, shifting criteria, or offering ad hoc justifications to maintain a desired narrative.



\textit{``This is not a bailout, this is considering providing certain things for certain industries. Airlines, hotels, cruise lines.''}

This type of bias could be considered a practical example of motivated reasoning. Unlike confirmation bias, which is mainly a subconscious phenomenon (and plays a role in many of the media biases discussed here), motivated reasoning is an active process in which one finds arguments to justify why one's dearly held beliefs and assumptions remain true in light of new information \citep{kunda1990motivated}.

In practice, this effect often reveals itself in the perception of political scandals. Behavior that would generally be universally condemned and viewed negatively, as expected when performed by a politician from an opposing political alignment, is often excused, relativized, and justified when committed by a politician with whom one identifies \citep{Lee2023}.

\subsubsection{Side Note Bias}

This type of bias arises when potential unnecessary and unrelated information and remarks are inserted in a discussion to advance a narrative or divert attention from the actual story.

\textit{``Meanwhile, ScotRail, under public ownership since April 2022, has paid nearly £2.5 million in compensation to passengers due to delays and cancellations, with the highest compensations issued for delays over 120 minutes.''}

Regardless of the actual subject under discussion, it's always possible to remind the audience of the preferred narrative, be it by a quick sideswipe or by bringing up marginally, if at all, related talking points. Even when conflicting information must be acknowledged, it can be softened or diluted by briefly circling back to the intended message.

Similar to the anchoring effect, introducing minor but relevant information before addressing the “bigger picture” can prime recipients to use the earlier detail as a reference point and thereby influence their judgments. For example, one experiment found that when participants were asked about their satisfaction with their dating life before reporting their overall life satisfaction, their overall ratings became highly correlated with their dating evaluations, a relationship that disappeared when the question order was reversed \citep{Strack1988PrimingCommunication}

Dilution can also occur by overwhelming the audience with an excessive amount of accompanying information, where the sheer volume overshadows the content, making it difficult to focus on the information meant to be suppressed. This strategy, especially with the focus on accompanying one controversy with a dozen others, has been popularized as ``flooding the zone'' among right-wing political strategists \citep{illing2020flood}.

A 2020 case study showed this principle at work during the Mueller investigation about potential Russian influence on the election of the then-sitting president: increased coverage of the investigation was followed by the president tweeting increasingly about unrelated issues, which in turn managed to shift the coverage from mainstream news media away from the topic \citep{lewandowsky2020using}.



\subsubsection{Whataboutism Bias}

This is the practice of responding to an accusation or the pointing out of a problem by making a counter-accusation or raising a different issue, without directly addressing the original argument.

\textit{``The media coverage of the Kavanaugh confirmation circus shows its strength in pushing a narrative, but sadly, journalists have failed to use their power to unite the world in opposition to China's sickening persecution of Uighurs and Hong Kong citizen.''}

While the specific term 'whataboutism' originated during the Troubles in Northern Ireland and has been frequently associated with Soviet and Russian propaganda, the strategy itself is way older, predating the Cold War, and has been widely used by various actors in international relations, especially in discussions of human rights abuses \citep{Dykstra2020}.

Its effectiveness in this field has, in the meantime, also been confirmed by research. In a study, the level of criticism for blatantly bad actions committed by a state, such as the mistreatment of refugees or interference in foreign elections, and the support for the USA taking action against these practices, could be reduced by the criticized state pointing out previous wrongdoing by the USA in similar matters. The more recent the misdeeds, the more effective this strategy was. Furthermore, in these experiments at least, the identity of the state responding with the counter-accusation did not matter, and the US counter-messaging did not prove to have a significant effect \citep{Chow_Levin_2024}.



\subsection{Dividing}

The bias types under this label are united in their efforts to sow division and turn people against each other.

\subsubsection{Discriminatory Bias}

This form of bias occurs when stereotypes, generalized or prejudiced statements and unequal representation are promoted or reiterated, reinforcing discrimination or biases against certain individuals or groups, often based on ethnicity, culture, nationality, social background, gender, sexual orientation or religious beliefs. 

\textit{``Women are naturally meant to be homemakers.''}

Because stereotypes serve as unavoidable mental shortcuts in how we perceive and classify others, it is easy for them to slip into reporting. Even when unintentional, their subsequent presence in the media serves to further normalize and solidify them within society, with the influence of media on group perception being the stronger the less real contact a person has with members of the stereotyped group \citep{Schiappa}.
At the same time, stereotypes shape not only how others perceive a group but also how members of that group view themselves, acting as a self-fulfilling prophecies \citep{Madon}.



\subsubsection{Gatekeeping Bias}

This refers to the attempt to define who legitimately belongs to a group or an identity or who may legitimately speak on a topic, and who does not, based on arbitrary and highly subjective criteria that lack objective justification.

\textit{``The only real union is the CWA at the Derby plant.''}


Research indicates that when people feel their group identity is challenged, they become more open to messages that reinforce what it allegedly means to be a 'true' member of the group \citep{white2018}. 
In one study, for example, men who were made to feel insecure about their masculinity by made up information about an alleged low score on different kinds of arbitrary tests, reacted by avoiding stereotypically feminine behavior and exaggerating their own stereotypical masculine characteristics \citep{Cheryan2015}.

Simultaneously, a perceived threat can lead people to raise the bar for what it means to belong to a group, not just for themselves, but for others as well. This can produce paradoxical outcomes. One study found that among individuals who feel threatened by immigration, gatekeeping attitudes are actually stricter in countries with relatively liberal integration policies, precisely because such policies make it easier for immigrants to establish themselves as part of their new home. In other words, when the criteria become easier to meet, those who see outsiders as a threat respond by simply raising the standards to prevent them of ever being part \citep{Uysal16032025}.

This demonstrates that, while gatekeeping is theoretically possible without constructing an adversarial “us vs. them” narrative, in practice it often becomes the first step toward doing exactly that.

\subsubsection{Us vs Them Bias}

This form of bias arises when humans are divided and assigned membership into ostensibly adversarial groups pitched against each other, an in-group, which is collectively and consistently portrayed positively, with legitimate motives and justified actions, and an othered out-group, which is consistently described with negative characteristics, illegitimate motives, and unjust actions.

\textit{``When they let, I think the real number is 15, 16 million people into our country, when they do that, we got a lot of work to do. They’re poisoning the blood of our country. That’s what they’ve done.''}

%
%
This is arguably the most historically harmful type of bias. No war, no genocide, no mass atrocities would even be conceivable if it weren't possible to convince a group of people that there is an ethnically, religiously, culturally, politically, or otherwise defined line between them and another group of people, and that everyone outside of this line is ``not like us.'' ``Not like us,'' the righteous ones, who are always only defending themselves against the threat posed by the Others, who, in contrast, pursue base motives and are morally inferior to our group in every way.

It appears that, as proposed by social identity theory, the desire to form groups and strengthen one's group identity by distinguishing and demeaning other groups \citep{Lalonde2002} is so deeply ingrained in humanity that even the most minor, insignificant, or random factors, such as preference for one painter over another, overestimating or underestimating the number of dots \citep{Tajfel1970}, or flipping a coin to determine heads or tails \citep{Deschrijver2025}, can be used to construct a group identity. This happens even without interaction with other members of the in-group or out-group, and results in preferential treatment for one's own group over the other.

Given this apparent tendency for group-based thinking, it is no surprise that there is evidence that, as with other biased media content, audience demand can be an important driver behind the tendency of newspapers to report a story through the lens of group identities \citep{Hopkins2025}.

Of course, news media not only plays a role in shaping which features group identities are constructed around, but also, in the absence of direct contact with members of a group, one's image and perception of that group is largely formed by its portrayal in the consumed media and by the representation of its members. Just as positive ``real-world'' contact with group members can improve related attitudes, while negative contact can worsen them, parasocial contact via media, whether through news or entertainment, can shape attitudes toward a group in both directions \citep{Banas2020}.

Given that a negative or outright hostile perception of a group is linked with a lesser desire to have contact with its members \citep{Croucher2017}, often institutionalized by policies like segregation or the closing of borders, negative media portrayal of the other group can be seen as a self-amplifying process, until the image of it is completely removed from the actual humans making up this group and only exists as an artificially made-up construct on which the most negative narratives can be projected.

The final, most devastating consequence of pushing an us-versus-them narrative in media can be physical violence. Studies have established a link between exposure to this kind of propaganda and increased participation in the genocidal murder of neighbors now deemed ``the enemy" in Germany \citep{Adena2015} and Rwanda \citep{Yanagizawa-Drott2014}, respectively.
It is not without reason that, after both genocides, the main propagandists pushing the dehumanization of the victims, without which those crimes would not have been possible in the first place, were put on trial, like others directly ordering and committing acts of violence \citep{wilson2015inciting}.
As Aldous Huxley once put it, ``the purpose of propaganda is to make one set of people forget that other sets of people are human." \citep{huxley1936olive}.
However, evidence from post-genocide Rwanda further shows that media campaigns and propaganda can also work toward dissolving group separation altogether, lowering the salience of ethnicity in one’s identity, and shifting the focus to what unites two sets of people rather than what separates them \citep{Blouin2019}.

\subsection{Misreasoning}

The bias in this category stems directly from logical errors in arguments, which can occur even when made in good faith.

\subsubsection{Causal Misunderstanding Bias}

This is when a cause-and-effect relationship between two variables is misunderstood or 
assumed without sufficient evidence or considering other factors.

\textit{``Now the [nuclear] power plant has been shut down. And Mrs. Grossmann, whom you probably also know, has now stated: ``We will shut down production for days and weeks at a time because we can no longer get electricity that we can afford.''}

When two events frequently occur together or in succession, such a relationship is often intuitively inferred without actual evidence to support it. However, the correlation might be an ``optical illusion'' in the first place, and even if it is not, that does not necessarily mean that one causes the other. It could simply be coincidence, or both events might share a common but unidentified cause. Additionally, while a cause-and-effect relationship may exist, presenting it as the sole or primary cause can be a significant oversimplification. Interestingly, when presented with the same data, people come to very different conclusions about cause and effect. Positive outcomes are far more easily attributed to policies aligning with one's own preferences, while negative outcomes are more quickly associated with policies one did not like in the first place.\citep{blanco2018}

Besides genuine misunderstandings, such false relationships can also be deliberately constructed to associate two issues with each other.

\subsubsection{Circular Reasoning Bias}

This occurs when the conclusion of a statement or argument is used as its own 
justification, essentially bypassing the requirement for evidence or logical reasoning.

\subsubsection{Burden Of Proof Bias}

This refers to the claim that something is true or false just because it has not yet been proven otherwise, often shifting the responsibility for evidence onto others and appearing in situations where getting absolute proof is unrealistic.

\textit{``The reason why people still question Obama's citizen status is one-fold: ``President Transparency'' has refused to release any original documents on the matter. He can end the controversy in a day by releasing original documents, but for some inexplicable reason he refuses, and his love-struck media never asks him why he won't.''}



Brandolini’s law states that “the amount of energy needed to refute nonsense is an order of magnitude greater than to produce it.” \citep{williamson2016take}.
The same dynamic appears here: it’s easy for someone to claim, for example, that the government is secretly run by lizard people, but proving that it isn’t is far more difficult, if not impossible. This asymmetry offers a kind of “protection” for such claims, because the person making them can pretend they’ve done their part and shift the burden of proof onto others to disprove them.

The reverse case, assuming something is false simply because it hasn’t yet been proven, can also be problematic, even its usually fairer. A lack of evidence can be a clue, but the circumstances under which this lack of evidence has been observed are often ignored. Consider the “black swan,” which was literally a metaphor for the impossible until actual black swans were discovered in Australia, or the giant squid, long dismissed as myth until physical evidence emerged in the 19th century.

This asymmetry is a challenge in science communication: explaining how strong a falsification actually is when rejecting a null hypothesis, and clarifying whether the result truly rules something out or whether certain limitations might weaken its power. 

In experiments, it could be shown that while generally positive evidence for something is more convincing than the absence of evidence, the absence of evidence still showed significant effects, especially in cases where there was a high number of attempts to find evidence, and when prior beliefs were already strong \citep{oaksford2004bayesian}.

\textit{``Cannabis is banned because it is an illegal drug.''}

Circular reasoning is a deceptive way to create a seemingly logical argument, where in reality, there is none. However, since the entire argument relies on the pre-existing acceptance of its premise, its persuasive power may not be very strong. In fact, research shows that around the age of 10, children tend to start finding circular arguments less convincing compared to actual explanations \citep{baum2008}.


\subsubsection{Generalization Bias}

This type of bias involves attributing actions, beliefs, goals or characteristics of individual members of to an entire group or institution, or conversely, assuming that individuals must share traits associated with their group.

\textit{``I know the mentality of people these days, it looks like this: Working from home, nobody wants to work anymore, nobody wants to carry boxes and the mobile phone is on silent mode from 9 pm. That might be the new generation, but it's not me.''}

This type of bias is closely linked to the concept of the representativeness heuristic, a mental shortcut in which people judge the probability or frequency of an event based on how much it resembles the typical case or stereotype they have in mind.

In a famous experiment, participants were presented with a description of a woman named Linda, who was described as being interested in social causes. They were then asked to judge which statement was more likely to be true: Linda is a bank teller or Linda is a bank teller and active in the feminist movement. A majority of participants stated that they considered the second option more likely, even though this cannot be true mathematically. Adding an additional condition can only decrease the likelihood (or keep it the same) \citep{Tversky1983}.

The reason for this bias is that people intuitively assume that if Linda is part of one group (people interested in social causes), she must also conform to the stereotypes associated with that group.

Real-world consequences of this kind of generalization include the increased likelihood that people belonging to a group associated with statistically higher crime rates, whether rightly or wrongly, may be found guilty of crimes more often than those who do not belong to such groups \citep{Curley2022}.

In the other direction, this also opens an avenue for shaping the perception of entire organizations, institutions, or movements by singling out individual voices or actions and presenting them as representative of the group. In fact, studies have shown that most Americans significantly overestimate the prevalence of ``party-stereotypical'' groups within both major political parties. This effect is most pronounced among those who claim to follow politics closely, suggesting a link to generalization in political coverage \citep{ahler2018parties}.



\subsection{Personalizing}

This group deals with types of bias that shift the discussion from a factual level to a personal level.

\subsubsection{Ad Hominem Bias}

This bias is when an argument is attacked by targeting the character, motives, or other attributes of the one making the argument, rather than addressing the substance of the argument itself.

\textit{``Any climate change message from somebody who owns a car is pure hypocrisy and should be completely ignored.''}

There are a number of reasons why one might choose not to listen to someone, whether due to past misjudgments, an alleged general weakness of character, or, very commonly, the accusation of hypocrisy.

In a study that examined the impact of different types of ad hominem attacks on the credibility of scientific findings  \citep{10.1371/journal.pone.0192025}, the accusations of conflicts of interest and past misconduct were particularly effective at significantly reducing credibility.

Part of the strength of this type of argument likely arises from the fact that it is not inherently illogical. Conflicts of interest can indeed be a problem, and as a proverb aptly puts it: ``A liar is never believed, even when he's telling the truth.''

On the other hand, one could argue that the strength of the ad hominem bias partly stems from the fact that, similar to other types of bias \citep{stager2006false}, it offers a way to avoid cognitive dissonance. Rather than engaging with information that contradicts one's worldview, one does not have to confront the argument at all.

To be distinguished from other personal attacks (see Mud \& Honey Bias), ad hominem bias is primarily characterized by the fact that the ultimate goal is still to discredit a specific argument or a particular piece of information by referencing its origin, whereas personal attacks aim to discredit the person (or organization) itself, independently of any specific argument.

\subsubsection{Association Bias}

This bias occurs when one links an individual, group, or policy to others with a strong positive or negative reputation by by overstating a tenuous connection or citing an irrelevant association.

\textit{``Football supports Trump in its promotion of racial division, the crushing of dissent, and the spread of misinformation, inequality, and brutality.''}








As indicated by the old saying ``birds of a feather flock together'', people tend, right or wrongfully, to assume those who associate with each other share similar traits. 
So, there exists a reputation spillover effect: people associated with others of high reputation and status will see their own perception boosted, while those associated with people viewed negatively will also lose social prestige \citep{Overton2021}.

Therefore, finding an angle that links a target to something with a reputation in the direction one want the target's image to move, be it actual contact, similarities between ideas or any other kind of perceived connection, might be an effective strategy towards achieving this goal.

Of course, this has to be differentiated from pointing out actual, deeper connections as part of a considerate analysis.



\subsubsection{Horse Race Bias}

This refers to the exaggerated portrayal of events and issues as personal wins or losses of an individual or a group, often as part of a competition between two or more parties.  

\textit{``Taylor Swift Scores Win Over Donald Trump.''}


This kind of reporting is not limited to, but is unsurprisingly especially prominent during, election season. During the US presidential elections in 2016, for instance, the majority of all election coverage could be classified as horse-race journalism, with only 10\% discussing the policy positions of the different candidates \citep{patterson2016}. The numbers in 2020 were not much different \citep{patterson2020}.

This carries some negative implications for the democratic process. First of all, it has been shown that, in order to stick to the competition frame, news reports tend to leave out candidates who are seen as ``no real'' competition \citep{Kirch2022}. This potentially contributes to a self-fulfilling prophecy, where those candidates fall short of their theoretical potential because they do not get the chance to present themselves and their policies to voters.

This is not the only way horse race reporting can influence what it claims to merely cover, as how close ``the race'' is portrayed might also affect turnout. Reporting as if a candidate has already won could paradoxically have opposite effects, such as lowering turnout due to complacency \citep{Westwood2020} or increasing it through bandwagon motivation \citep{MORTON201565}. 

Though there is debate over which of these effects is more common and how strong they are, some long-term consequences appear to be more widely agreed upon. These include a general increase in political cynicism and a decrease in political knowledge \citep{Zoizner2021}.

Once again, this type of bias appears to be driven largely by consumer demand, one study of Portuguese election coverage showed that portraying elections in the discussed manner led to higher reader engagement \citep{Gonçalves2022}.

\subsubsection{Mud \& Honey Bias}

This is the practice of using personal attacks, rumors or unfounded allegations to damage the reputation of an individual or a group, or the opposite tendency to excessively praise or 
idealize them without regard for objective evaluation or criticism. 

\textit{``He’s a 78-year-old white male millionaire seeking the affections of the wokesters, a gruff socialist with bedraggled hair and terrifying policy proposals.''}

Findings on the effectiveness of this tactic have been mixed. In classical negative political campaigning, while it can successfully lower a candidate’s vote share, either by persuading their voters to choose differently or discouraging them from voting altogether, it may also backfire on the attacker, reducing their own support, making third parties, which are not part of this (at least in multi-party systems), the true beneficiary \citep{Galasso2020Positive}.

Therefore, it might be a good strategy for a party to let the media do the dirty work in order to avoid the backfiring, as both positive and negative portrayals of a candidate still influence popularity (even though there is no agreement on whether positive or negative coverage has a greater effect) \citep{Annelien2024}.

Further research into the specific variation of establishing degrading nicknames for political opponents, which has been described as the attempt on a "transformation of individuals into stereotypes" by Timothy Snyder \citep{snyder2017tyranny}, indicates that such tactics are, as so often with propaganda, mainly effective in deepening attitudes among recipients, which already have a negative opinion of the politician \citep{johnson2021sleepy}.

\subsection{Preferring}

This group focuses on bias types that revolve around the explicit expression of support for particular subjective beliefs or opinions.

\subsubsection{Commercial Bias}

This refers to the practice of emphasizing or directly promoting certain companies, their products and services, as well as the support of favorable narratives due to underlying commercial interest. 

\textit{``At the Reuters’ Global Energy Transition 2024, Chevron Vice President of Strategy \& Sustainability Molly Laegeler discussed how to build a resilient strategy and incorporate it into everything your company does as well as how her company leans on its strengths to advance lower carbon solutions.''} 

With the exception of news media supported by not (directly) profit orientated external funding, such as state-funded outlets, publicly financed organizations, or those backed by wealthy patrons, news organizations face pressure to turn a profit or at least break even to stay afloat. Besides subscription fees and newsstand sales, advertising has traditionally played a major role in keeping the lights on. While news organizations typically maintain a theoretical separation between advertising and editorial content, the reality is more complex.

For advertisers, the value of placing ads in a reputable newspaper extends beyond its large audience, it also derives from the newspaper’s credibility and the trust it holds with its readers. Therefore, a complete separation between editorial and advertising content cannot realistically exist, as the former serves as the foundation for the latter. 

Therein lies a conflict of interest: if editorial decisions are not well received by an advertising client, they might have been a client for the longest time. In fact, several studies have shown a link between advertisers in a news outlet and how that outlet reports on certain issues. One study, for instance, demonstrated that between 2000 and 2014, US newspapers reported significantly less on major safety recalls affecting car models from brands that were frequent advertisers \citep{Beattie}.

Beyond decisions directly relating to a company or product, this type of bias can also manifest in the way broader developments and debates are reported, especially when the topic is relevant to a company's profits. For example, the scientific discourse around global warming, its causes, and efforts to mitigate it directly impacts fossil fuel-dependent companies. There is evidence suggesting that newspapers have reported more skeptically on climate change when anticipating increased advertising from car manufacturers \citep{BEATTIE2020104219}.

For the sake of completeness, it should also be noted that commercial bias is not solely driven by advertising. Privately owned media outlets may also be partially owned by companies or individuals with other commercial interests, which could influence editorial decisions. Additionally, advertising as a means of indirectly shaping media coverage is not exclusive to private entities, governments have also been known to use it for similar purposes \citep{doi:10.1177/19401612241285672}.

\subsubsection{Ideological Bias}

This refers to an inclination toward a specific ideology, belief system, or worldview, be it economic, political, religious, or similar frameworks, or associated individuals, organizations, and symbols, resulting in inherent favoritism for aligned ideas while disregarding or disparaging opposing perspectives.

\textit{``The Cuban government enjoys wide popular support in large part because of the genuine leadership provided by that the Cuban Communist Party. Far from being an exploiting ruling class, its leadership is based on deep connections with the Cuban working class.''}

Growing up and living in a world far too complex to fully grasp without mental shortcuts, and in societies shaped by a variety of ideological frameworks, whether cultural, political, religious, or economic, no one can genuinely claim to possess a worldview untouched by ideological influence.

In fact, one could argue that ideology is the ultimate stereotype, providing a comprehensive guide which creates the illusion of understanding the world.

Yet it remains common for people to see their own views as “common sense,” as if those beliefs were the result of pure, unbiased reason and an objective reading of the world as it truly is, while assuming that those who disagree must be blinded by ideology (a phenomena usually referred to as naive realism) \citep{pronin2002bias}.




\subsubsection{Opinionated Bias}

This refers to the inclusion of subjective opinions, beliefs, or interpretations portrayed as objective, factual reporting, obscuring the line between verifiable evidence and personal perspective.

\textit{``And we know that the promises and rosy visions of the future made by those responsible in the Harz National Park are not realistic.''}

Naturally, regardless of the specific type, media bias is always a form of expressing opinion, after all, what is an “existing narrative” if not a subjective interpretation of the world? This type of bias, however, is specifically characterized by the fact that opinion is communicated quite openly, without taking the detour of packaging it in a particular rhetorical form.
If this weren’t usually accompanied by presenting the opinion as fact, one could almost call it the most honest form of bias.

While opinion pieces are a common element of modern journalism, audiences sometimes struggle with differentiating between opinion and fact. In a Pew Research Center study, only 50\% of participants were able to correctly classify at least 4 out of 5 factual statements, with the number being 59\% for classifying opinions \citep{mitchell2018distinguishing}. Interestingly, those who were better at differentiating between the two also more often had a higher general trust in the media. Furthermore, both Republican- and Democratic-leaning participants were more likely to rate a statement as factual when it appealed to their side, while facts not fitting their worldview were often incorrectly labeled as opinionated, showing the essence of this bias type: One is often so confident in their opinion that they no longer consider it an opinion, but a fact.


\subsection{Framing}

The types in this group are all about presentation, (miss)representing topics or arguments in a way that supports a specific reading.

\subsubsection{Emotional Sensationalism Bias}

This is when sentences use alarming, baiting, hyperbolic, or provocative language designed to evoke (strong) emotions, while often focusing predominantly on negative events, 
aspects or interpretations.

\textit{``Even the EU is shocked by France's move against the USA!''}

This bias relates to the cognitive principle of negativity bias, which suggests that people naturally give greater attention to negative experiences, even when positive ones are more abundant. This is often explained by a presumed evolutionary advantage of focusing on potential dangers \citep{Norris2019}.

However, in today’s modern world, where news media provide us with a 24/7 window into tragedies, crises, and catastrophes happening worldwide, over which we often have no direct agency, this mechanism offers little advantage. Still, studies show that news content focusing on negative events or presented in a way that arouses negative emotions is clicked on much more often than neutral or positive news. In turn, this creates a financial incentive for news companies to prioritize such content, making negativity bias one of the most prevalent biases in media \citep{Zhang2024}.

This is problematic because it leads to a distorted perception of reality, one that is often seen as more hopeless, bleak, and dangerous than it actually is. This has far-reaching consequences for political debates and policymaking, which may end up being shaped around a world that, in reality, does not exist. At the same time, this sense of helplessness can cause people to withdraw into their private lives and disengage from efforts to improve society altogether \citep{Galpin2017}.

Beyond its societal consequences, this bias is also linked to individual negative effects on mental health. ``Doomscrolling", the excessive consumption of large quantities of negative news online which is especially prevalent among young people, has been associated with increased anxiety, depression, and feelings of isolation \citep{Rodrigues2022}.

In response to these concerns, recent years have witnessed the emergence of several news outlets as part of a counter movement, dedicated solely to providing positive and solution-oriented stories \citep{goodnews}.

\subsubsection{Empty Symbol Bias}

This type of bias describes the invoking of symbols, such as phrases, slogans, concepts, ideals, historical events, or public figures, which are charged with meaning beyond their objective reality and usually broad enough to invite personal projection.

\textit{``National Day is a lofty and inspiring national occasion that embodies the spirit of unity, cohesion and belonging among citizens, and confirms their pride in the memory of the founder of the state, Sheikh Jassim bin Mohammed bin Thani, who laid the foundations of a strong and unified state, and made Qatar a model of steadfastness and independence.''}

In Public Opinion, Walter Lippmann stresses the importance of symbols for those who aim to control the ``approaches of public policy'' as they ``secure unity and flexibility without real consent'', not standing for ``specific ideas, but for a sort of truce or junction between ideas''. ``And as long as a particular symbol has the power of coalition, ambitious factions will fight for possession'', he further notes \citep{lippmann1922public}.

The importance of being perceived as possessing a certain symbol has, in the meantime, been confirmed by contemporary research. There is evidence, for instance, that in the USA, it is Republican candidates who benefit from an increased presence of national flags in election campaigns, even when it is a Democratic candidate who displays the flag. Unsurprisingly, perhaps, this effect is strongest among voters who consider themselves especially patriotic \citep{Kalmoe2016CueingPP} .

Interestingly, the influence of being exposed to symbols, specifically national flags, in shaping opinion has been demonstrated, even when the exposure was too short to be consciously recognized \citep{Ran2007}.

However, the meaning of a symbol is not uniform across all groups that might ascribe meaning to it. When, for one group, the symbol is seen as belonging to an out-group, it might even evoke completely different, even negative, associations. Symbols of one’s own group overall still appear to facilitate a stronger emotional response, though \citep{muldoon2020flagging}.

\subsubsection{False Balance Bias}

This type of bias occurs when 
opposing viewpoints are presented as equally credible or significant, despite a clear consensus or evidence favoring one side. 

\textit{``Barack Obama says he was born in Hawaii, and since no one has shown any proof he was born in Kenya or elsewhere, it's OK to conclude he was born in Hawaii. [...] It's OK though for others not to use my deferential standard and continue to question whether Obama was born in Hawaii.''}

False Balance ironically often stems from the fair and noble journalistic aspiration to avoid bias by presenting both sides of a debate, allowing audiences to form their own opinions. The problem arises when this principle is applied not just to opinions or interpretations, where no objectively ``correct" answer exists, but also to facts, where the truth does not lie somewhere in the middle and is objectively verifiable. 

This way, journalistic ethics can be weaponized by interest groups to sow doubt and stall political or societal action, because, as one tobacco industry executive once put it, ``doubt is the best means of competing with the body of fact” \citep{michaels2008doubt}.
%
\footnote{Interestingly, many PR specialists originally hired by the tobacco industry for this purpose later went on to work for fossil fuel companies, helping to cast doubt on the existence and causes of global warming, arguably the most well-known example of False Balance \citep{ciel2016oil}.}
%
Even when fact-checked, merely including verifiably false positions can sway the audience toward the intended false narrative \citep{barrera2020facts}.
A key factor enabling this tactic is that explicitly accusing journalists of bias, whether the accusation is true or not, can pressure them into ``over-correcting" their reporting. In an effort to prove their neutrality, they may deliberately integrate talking points from the opposing side, even when those points lack factual basis \citep{panievsky2022strategic}.

In her critique, Sarah Stein Lubrano identifies false balance as a symptom of a deeper problem: in modern democracies, pitting two opposing sides against each other in debate is often treated as the cornerstone of political opinion-forming, operating under the assumption that stronger arguments will naturally reveal the truth. This, she argues, is a false assumption. Debates, rather than resolving conflicts, often reinforce them, preventing a more systemic examination of the issues. They can mask the reality that many conflicts are driven less by competing arguments than by clashing interests, and can even serve as a form of coercion, granting those in power legitimacy to pursue their agenda under the guise that “everyone has had their say.” \citep{lubrano2025dont}.

\subsubsection{False Dichotomy Bias}

This bias occurs when a complex issue is presented as having only two opposing alternatives or being without alternative altogether, even though there might be more possible solutions, positions or outcomes.

\textit{``Society must choose: embrace advanced AI with potential risks or stifle innovation and fall behind technologically.''}

False dilemmas can be used to push talking points and force decisions that might otherwise be seen as too unpopular or radical. This is done by contrasting them with an even more undesirable alleged alternative, thereby presenting them as the lesser evil or a necessary choice. The narrative goes: ``If you don't support A, then you support B, and you certainly don’t want that. So A is the better option." Framing the support of one option mainly as the rejection of an unpopular option, can be highly effective, as demonstrated by a study aptly titled ``Rejecting a Bad Option Feels Like Choosing a Good One." \citep{Perfecto}

\subsubsection{Flawed Comparison Bias}

This bias is characterized by drawing analogies or comparisons between two or more things things that may share superficial similarities but are ultimately fundamentally different.


\textit{``One should have just asked a Swabian housewife: You can't live beyond your means in the long run. That is the core of the [financial] crisis.''}

It can be an oversimplification of complex issues by comparing them to something much simpler often invokes images that are already oversimplified stereotypes, and can lead to practical problems when reality turns out to be more complicated then the stereotype.

As for persuasiveness, there is evidence that analogies from daily life can enhance support for abstract policies when paired with a clear supporting rationale, a combination that proves more persuasive than using either strategy on its own \citep{Barabas2020}.

There are also scenarios where analogies did not do anything at all to influence opinions, instead the support of the analogy was entirely dependent on pre-existing support for the policy, not the other way around \citep{Barnes_Hicks_2022}.

One key factor, as with other types of persuasion attempts, may be how invested people are in their opinion on a matter in the first place and how much they think they know about it \citep{petty2002thought}.

This bias can further be exploited to discredit an idea or action by linking it with something universally condemned. Conversely, it can be used to downplay a negative issue by drawing a false equivalence with something bad but less severe.


%
%
%
%



\subsubsection{Magnitude Bias} 

This type of bias arises when the severity or relevance of an issue is relativized, downplayed or exaggerated. 

\textit{``The worst bureaucracy law of all time!''}

No issue presented in the news media exists in a vacuum, audiences naturally seek to understand how significant it is relative to others.

And outright telling them might be a surprisingly effective strategy, as shown in advertising \citep{Stern02012020} or testimony \citep{DESAI2021143} where the positive effects of hyperbole in particular on persuasion have been demonstrated. For skeptical individuals, however, such claims can backfire, especially when their initial convictions do not align with the premise of the hyperbole, and can actually reduce credibility, thereby widening the gap between “the believers” and “the skeptics” \citep{Boeynaems16032021}.

While hyperbole often relies on presenting an issue without a concrete point of reference, relativizing or downplaying can exploit the cognitive bias known as anchoring. In this phenomenon, people’s judgments are subconsciously influenced by an initial piece of information. It has been shown, for example, that the number people suggest when asked how many immigrants their country should accept is heavily influenced by whatever number they were previously exposed to in a fictional political proposal. When the proposal includes a higher number, participants tend to recommend a higher number themselves \citep{Arceneaux2024}.
(Interestingly, similar effects were also demonstrated in scenarios where the anchor was completely unrelated and randomly generated) \citep{Ariely2003}.

\subsubsection{Normalwashing Bias}

This bias involves the normalization or reputation laundering of organizations, individuals, or circumstances by avoiding describing them with the objectively applicable but potentially contentious descriptions and context.

\textit{``In remarks about migrants, Donald Trump invoked his long-held fascination with genes and genetics.''}

Similar to False Balance, this bias can paradoxically stem from an overly strong attempt to appear neutral and impartial. Gaye Tuchman once described the journalistic notion of objectivity as a ``strategic ritual protecting newspapermen from the risks of their trade'' \citep{tuchman1972objectivity}.

In this context, a journalist might find it safer to label an issue as ``controversial'' rather than assessing which side of the debate is more substantiated. However, when a politician makes a statement that is objectively racist according to socially accepted definitions, refusing to call it as such constitutes a bias in itself. Likewise, uncritically celebrating an actor’s seemingly positive action, without situating it within a broader context, can produce a skewed narrative, effectively “washing” (a term often used with different prefixes to describe various manifestations of this phenomenon) their behavior and presenting a distorted image. 

Exemplary for “greenwashing,” empirical evidence suggests that such tactics are effective when perceived as authentic by the general public but tend to backfire when their authenticity is questioned \citep{Persakis2025}. This underscores the crucial role that journalists, as influencers of public opinion, play in determining the success of these tactics.

In general, it has also been argued that a practice of objectivity and neutrality centered on avoiding controversy and alienating audiences primarily benefits those already influential in society. This approach reinforces their views as ``normal'' and the status quo, while marginalizing voices that are not part of the mainstream conversation and those, who question the status quo \citep{Hampton2008Objectivity}. 

In this sense, even genuinely well-intentioned reporting on real acts of charity and individual triumph over adversity can serve to reinforce the very conditions it implicitly agrees are undesirable. By failing to ask why these circumstances exist in the first place, and instead presenting them as a kind of “normal reality,” such reporting normalizes systemic injustice. Just as many of the ``washing''-terms have been introduced rather recently, this dynamic has been captured by the contemporary online-coined satirical metaphor of the “orphan-crushing machine”, used to describe stories that celebrate raising money to save orphans from being crushed by the metaphorical machine, yet never pause to ask why such a machine exists in the first place \citep{WiktionaryOrphanCrushingMachine}.

\subsubsection{Rhetorical Bias}

This refers to the use of rhetorical devices, such as repetition, sarcasm, rhyme, irony, or humor, in discussing an issue to shape its perception.

\textit{``At the time of publication, Christian Lindner was last seen on the deck of the burning ship, leaning over a burnt-out Porsche, crying loudly and vowing never to forget it.''}





The systematic use of rhetorical devices in persuasion dates back thousands of years. In ancient Greece and Rome for example, rhetoric was a core part of a lawyer’s education, because long before forensics and modern criminology could provide hard evidence, court cases often hinged on which side could better sway the judges \citep{Adeodato2019}.

Today, the persuasive effectiveness of numerous rhetorical techniques has been has confirmed by empirical research, such as the use of rhymes \citep{McGloneTofighbakhsh1999} and self-effacing humor \citep{Lyttle2001}.

The mode of presentation itself can also already make an issue appear more or less serious, the underlying severity of social issues presented in a humorous setting (like satire and political comedy) appear to be more easily discounted than when presented with a more serious tone \citep{Nabi2007}.

\subsubsection{Straw Man Bias}

This occurs when a position or argument is 
misrepresented and distorted in a way that makes it easier to attack or refute, often by oversimplifying or exaggerating it.

\textit{``All other ‘methods’ of informing people correctly have unfortunately been exhausted? The climate issue, a global issue with many unknowns, is no longer capable of any development and has been declared over? And a few activists are supposed to be the only ones who have the climate under control and now have to drive the whole world before them?''}


The misrepresentation of arguments opposing one's one view is actually quite common, some studies show that, even when given an incentive to be as accurate as possible in describing the position of ``the other side'', people often struggle to give a fair and nuanced retelling or even detect when a representation of an argument is actually a straw man \citep{Michael2022}.

Experiments indicated that personal relevance (i.e., how meaningful the subject is to the receiver) impacts the straw man's effectiveness in influencing opinion. The less motivation participants had to critically engage with an argument, the more effective the straw man was. Cognitive closure (see Speculation Bias) also appears to play a role, as participants with a high need for it were more easily influenced by this bias \citep{Bizer2009}.

There is also research suggesting that certain linguistic constructions can increase the effectiveness of straw man arguments, namely, attacking a specific argument or conclusion rather than the more fundamental standpoint behind it, avoiding logical connectors like ``because'' or ``since'' when linking two misrepresentations, and using an explicit misrepresentation that closely resembles the language of the original argument, rather than an indirect rephrasing \citep{SCHUMANN20191}.

\subsubsection{Word Choice Bias}

This type of bias arises when certain words and expressions with inherently positive or negative connotations, euphemisms, dysphemisms, or (strong) adjectives, adverbs and linguistic markers are chosen that influence perception and imply a judgment about a topic.

\textit{``Hordes of punks swarmed the island in the summer and caused chaos.''}

In many cases, connotated words have integrated so well into daily language that they are usually employed without a second thought to the narrative they support.
Several decades after most states decided to re-frame the official job title for the political role concerned with organized warfare as ``minister of defense'' \citep{Dinstein_2017}, for instance, this term and the narrative it entails are rarely questioned. 

This holds true even for articles naming and discussing wars of aggression waged by the respective country, where the title ``minister of defense'' is frequently used without any recognition of the inherent contradiction or mendacity \citep{dw2022russia}.


This, despite clear evidence that even the substitution of a single term can alter public opinion and political attitudes. For instance, after the Associated Press decided to retire the term ``illegal immigrant'' in favor of arguably less judgmental alternatives like ``undocumented immigrant'', individuals exposed to this change through AP-reliant local media showed significantly lower support for restrictive immigration policies \citep{djourelova2023persuasion}.

Even the outcome of the famous Prisoner's Dilemma game (where two players can either cooperate for mutual benefit or betray their partner for individual gain) can be changed simply by renaming it. When the game was called the ``Community Game,'' participants were up to twice as likely to cooperate as when it was named the ``Wall Street Game," after the metonym for the U.S. financial industry, which is often associated with high competition and a ``winner takes it all" attitude \citep{liberman2004name}.

Because the use of connotative words can occur so naturally in everyday language, it is not surprising that many analyses (including our own) find this to be the most prevalent form of bias \citep{RodrigoGines-CarrilloDeAlbornoz-Plaza:2024:ExpSysAppl}.

\section{Conclusion, Limitations and Future Work}

In this article, we introduced a sentence-level taxonomy of 38 media bias types, organized into functional groups. By focusing on linguistic and rhetorical techniques rather than ideological labels, we offer a practical, extensible framework to identify and analyze bias in news. 

Our taxonomy does not claim to be exhaustive (which is why we explicitly chose an extensible design). The practical examples used to develop this taxonomy are drawn primarily from the German-speaking world (with a focus on Germany) and the English-speaking world (with a focus on the USA), due to our linguistic and cultural background, as well as the availability of resources. These examples are a diverse mix from various sources, but ultimately, they represent only a sample from a vast media ecosystem. We cannot guarantee their representativeness for the individual language regions, let alone for those not covered.

Human language, at the end of the day, is not always unambiguous. It is complex, nuanced, ambiguous, and often subjective in its interpretation. Any taxonomy dealing with human language will, therefore, also be subject to ambiguity. Where exactly the line between objective and biased content, or between two types, lies, how they can be split, merged, or grouped, will always remain a matter of debate. While we are confident that our systematic approach provides satisfying answers to these questions, we acknowledge that other perspectives are equally legitimate. We are pleased to contribute to this ongoing discussion and will continue to refine our taxonomy based on feedback from researchers, media practitioners, and communication professionals, as well as insights from new experiments. Moving forward, we aim to deepen our investigation into the practical applications of our taxonomy by integrating it into automated detection efforts and exploring its potential for media literacy interventions and educational initiatives.

\end{doublespace}

\newpage

\section*{Acknowledgements}

We would like to thank our students and colleagues
who helped us test our news bias model and who
assisted in collecting many news articles, biased
and neutral. Special thanks to Katharina Weiß, Maximilian Schönau and Michael Reiche for their feedback. This research was funded by the
Hitech Agenda program of the Free State of 
Bavaria.



\bibliography{references}

\newpage

\section{Supplementary Material}

\begin{longtable}{p{3cm} p{4cm} p{4cm} p{4cm}}
\caption{Mapping: Ours → DaSanMartino et al. → Spinde et al. → Rodrigo-Ginés et al.}
\label{table:comparison} \\      

\hline
\textbf{Ours} & \textbf{DaSanMartino et al.} & \textbf{Spinde et al.} & \textbf{Rodrigo-Ginés et al.} \\
\hline
\endfirsthead

\multicolumn{4}{c}%
{{\bfseries \tablename\ \thetable{} -- continued from previous page}} \\
\hline
\textbf{Ours} & \textbf{DaSanMartino et al.} & \textbf{Spinde et al.} & \textbf{Rodrigo-Ginés et al.} \\
\hline
\endhead

\hline \multicolumn{4}{r}{{Continued on next page}} \\ \hline
\endfoot

\hline
\endlastfoot

Opinionated Bias & -- & Framing Bias; Epistemological Bias; Statement Bias & Opinion statements presented as facts bias; Subjective qualifying adjectives bias \\

Ideological Bias & Flag-waving; Slogans; Thought-terminating cliché & Framing Bias; Statement Bias & Slant bias \\

Commercial Bias & -- & Framing Bias; Statement Bias & Slant bias; Subjective qualifying adjectives bias \\

Vagueness Bias & -- & Epistemological Bias; Spin Bias & Omission of source attribution bias \\

Ad Hominem Bias & Name calling or labeling; Doubt & Linguistic Intergroup Bias; Framing Effects & Ad hominem/mudslingin bias \\

Mud \& Honey Bias & Name calling or labeling & Linguistic Intergroup Bias; Phrasing Bias; Hate Speech & Ad hominem/mudslingin bias \\

Horse Race Bias & -- & Framing Bias; Framing Effects & Flawed logic bias \\

Association Bias & Reductio ad hitlerum & Connotation Bias; Linguistic Intergroup Bias; Statement Bias; Framing Effects & -- \\

Suggestive Questioning Bias & Doubt & Epistemological Bias; Bias by Semantic Properties; Statement Bias; Framing Effects & -- \\

Projection Bias & -- & Epistemological Bias; Statement Bias; Framing Effects & Mind reading bias \\

Speculation Bias & -- & Epistemological Bias; Statement Bias; Framing Effects; Informational Bias & Flawed logic bias \\

Unsubstantiated Claims Bias & -- & Epistemological Bias; Statement Bias & Unsubstantiated claims bias \\

Burden of Proof Bias & -- & -- & Flawed logic bias \\

Circular Reasoning Bias & -- & -- & Flawed logic bias \\

Causal Misunderstanding Bias & Causal oversimplification & Framing Effects & Flawed logic bias \\

Generalization Bias & -- & -- & Flawed logic bias \\

Word Choice Bias & Loaded language & Framing Bias; Epistemological Bias; Connotation Bias; Linguistic Intergroup Bias; Statement Bias; Phrasing Bias; Hate Speech & Slant bias; Bias by labeling and word choice; Subjective qualifying adjectives bias \\

Emotional Sensationalism Bias & Appeal to fear/prejudice; Name calling or labeling & Framing Bias; Connotation Bias; Spin Bias; Phrasing Bias; Framing Effects & Sensationalism/emotionalism bias; Subjective qualifying adjectives bias \\

Magnitude Bias & Exaggeration or minimization & Framing Bias; Epistemological Bias; Connotation Bias; Spin Bias; Phrasing Bias; Framing Effects & Slant bias; Subjective qualifying adjectives bias \\

Rhetorical Bias & Repetition & Bias by Semantic Properties & -- \\

Us vs Them Bias & Appeal to fear/prejudice & Framing Bias; Connotation Bias; Linguistic Intergroup Bias; Phrasing Bias; Framing Effects; Hate Speech & Flawed logic bias \\

Discriminatory Bias & -- & Framing Bias; Connotation Bias; Linguistic Intergroup Bias; Phrasing Bias; Framing Effects; Hate Speech & -- \\

Gatekeeping Bias & -- & Framing Bias; Connotation Bias; Linguistic Intergroup Bias; Framing Effects & -- \\

Cherry Picking Bias & -- & Spin Bias; Framing Effects & Slant bias; Omission bias; Commission bias \\

Empty Symbol Bias & Flag-waving; Slogans; Thought-terminating cliché & Framing Bias; Phrasing Bias; Framing Effects & -- \\

False Dichotomy Bias & Black-and-white fallacy & Statement Bias; Framing Effects & Flawed logic bias \\

Flawed Comparison Bias & Reductio ad hitlerum & Connotation Bias; Statement Bias; Framing Effects & -- \\

Straw Man Bias & Straw man & Statement Bias; Framing Effects & Slant bias \\

Shifting Goalpost Bias & -- & Framing Effects & -- \\

Social Compliance Bias & Bandwagon; Appeal to authority & -- & Flawed logic bias \\

Source Selection Bias & -- & Spin Bias; Framing Effects & Slant bias; Commission bias; Source selection bias \\

Anecdotal Evidence Bias & -- & Spin Bias; Framing Effects & Slant bias; Flawed logic bias \\

Whataboutism Bias & Whataboutism & Epistemological Bias; Statement Bias & Flawed logic bias \\

Side Note Bias & Red herring & Spin Bias; Framing Effects & Flawed logic bias \\

No Discussion Bias & Dictatorship & Framing Effects & -- \\

Claim \& Blame Bias & Causal oversimplification & Bias by Semantic Properties; Statement Bias & -- \\

Normalwashing Bias & -- & Connotation Bias; Statement Bias; Spin Bias; Framing Effects & Subjective qualifying adjectives bias \\

False Balance Bias & -- & Statement Bias; Framing Effects & Flawed logic bias \\

Non-sentence level types & -- & Selection Bias; Coverage Bias; Proximity Bias; Selective Exposure; Partisan Bias; Group Bias & Bias by placement; Size allocation bias; Picture selection bias; Picture explanation bias \\

\hline
\end{longtable}

\newpage

\begin{table}[htbp] 
\centering
\caption{Prevalence of bias types (sample of 155 biased sentences)}
\label{tab:bias-absolute-percentages-multicol}
\begin{tabular}{@{} l l S[table-format=2.2] @{}}
\toprule
\textbf{Group \% } & \textbf{Bias type} & {\textbf{Type \% }} \\
\midrule

\multirow{10}{*}{Framing (68.39\%)} 
  & Word Choice Bias & 58.06 \\
  & Emotional Sensationalism Bias & 15.48 \\
  & Magnitude Bias & 8.39 \\
  & Rhetorical Bias & 7.10 \\
  & Empty Symbol Bias & 5.81 \\
  & False Dichotomy Bias & 4.52 \\
  & Flawed Comparison Bias & 3.87 \\
  & Straw Man Bias & 3.87 \\
  & Normalwashing Bias & 3.87 \\
  & False Balance Bias & 1.94 \\

\midrule

\multirow{5}{*}{Asserting (54.19\%)} 
  & Vagueness Bias & 24.52 \\
  & Unsubstantiated Claims Bias & 19.35 \\
  & Speculation Bias & 10.32 \\
  & Projection Bias & 9.03 \\
  & Suggestive Questioning Bias & 5.16 \\

\midrule

\multirow{3}{*}{Preferring (38.71\%)} 
  & Opinionated Bias & 36.77 \\
  & Ideological Bias & 9.03 \\
  & Commercial Bias & 1.29 \\

\midrule

\multirow{4}{*}{Personalizing (23.87\%)} 
  & Ad Hominem Bias & 7.74 \\
  & Mud \& Honey Bias & 7.10 \\
  & Association Bias & 7.10 \\
  & Horse Race Bias & 5.16 \\

\midrule

\multirow{4}{*}{Misreasoning (21.29\%)} 
  & Generalization Bias & 14.84 \\
  & Causal Misunderstanding Bias & 9.68 \\
  & Circular Reasoning Bias & 1.29 \\
  & Burden Of Proof Bias & 0.65 \\

\midrule

\multirow{5}{*}{Deflecting (18.71\%)} 
  & Claim \& Blame Bias & 7.10 \\
  & Side Note Bias & 5.16 \\
  & No Discussion Bias & 3.87 \\
  & Shifting Goalpost Bias & 1.94 \\
  & Whataboutism Bias & 1.94 \\

\midrule

\multirow{3}{*}{Dividing (16.13\%)} 
  & Us vs Them Bias & 10.32 \\
  & Discriminatory Bias & 5.16 \\
  & Gatekeeping Bias & 4.52 \\

\midrule

\multirow{4}{*}{Confirming (16.13\%)} 
  & Cherry Picking Bias & 5.81 \\
  & Social Compliance Bias & 5.81 \\
  & Source Selection Bias & 4.52 \\
  & Anecdotal Evidence Bias & 3.87 \\

\bottomrule
\end{tabular}
\end{table}

\end{document}